\documentclass{article} %
\usepackage{preprint,times}

\usepackage{amsmath,amsfonts,bm}

\def\eqref#1{equation~\ref{#1}}

\def\1{\bm{1}}

\DeclareMathAlphabet{\mathsfit}{\encodingdefault}{\sfdefault}{m}{sl}
\SetMathAlphabet{\mathsfit}{bold}{\encodingdefault}{\sfdefault}{bx}{n}

\usepackage{hyperref}
\hypersetup{colorlinks=true,linkcolor=[rgb]{0.10,0.25,0.55},citecolor=[rgb]{0.10,0.25,0.55},urlcolor=[rgb]{0.10,0.25,0.55}}
\usepackage{url}
\usepackage{graphicx}
\usepackage{amsthm}
\usepackage{booktabs}
\usepackage{algorithm}
\usepackage{algorithmic}
\usepackage{float}
\usepackage{xcolor}
\newtheorem{theorem}{Theorem}
\newtheorem{proposition}{Proposition}
\newtheorem*{theoremprecise}{Theorem~\ref{thm:char} (precise form)}

\title{Constrained Diffusion from Binary\\ Evaluations via Learned Doob $h$-Functions}

\author{Ssaumya Kammal Jaiswal \\
University of Pennsylvania
\And
Mahfuza Farooque \\
Pennsylvania State University
}

\usepackage{array}

\pprfinalcopy
\begin{document}

\maketitle

\begin{abstract}
Conditional generation under constraints that can only be checked on a finished sample, such as a molecular property filter, remains challenging for diffusion models: most guidance methods require differentiable measurement operators, paired conditional data, or rollouts of the base model at every step. We present a framework grounded in the Doob $h$-transform that reduces conditional generation to learning a single function, the probability that a noisy sample denoises into the constraint set. We learn this function from binary clean-sample evaluations alone, by squared-error regression on forward-noised labels; the base model is frozen, one lightweight head is trained per constraint, and noisy-state classifier guidance arises as a special case. The exact guidance step fades where the constraint signal provably vanishes and carries a factor of the inverse probability across samples, which normalized guidance discards: with the schedule and mean step size held fixed, equalizing the step costs 17.0 to 18.5 points of constraint satisfaction on CIFAR-10 at comparable FID, across independently trained estimators. Guidance is more than $50\times$ cheaper per satisfying sample than rejection sampling at a 1.7\% base rate, raises a Lipinski-type property filter on QM9 from 54.3\% to 96.8\%, and lifts satisfaction on Stable Diffusion 1.5 from 7.2\% to 60.0\% at one sixteenth of the cost of derivative-free selection, which reaches 9.0\%. These results demonstrate that binary evaluations of clean samples are sufficient supervision for constrained generation across pixel, latent, and molecular diffusion.
\end{abstract}

\section{Introduction}
 
Many constraints of practical interest can be verified by evaluating a program on a
completed sample but cannot be expressed as differentiable functions of it: whether a
molecule satisfies a property filter, whether a graph contains a required substructure,
or whether an image satisfies a pixel-level predicate. Any thresholded score defines
such a constraint, and filters of this kind are routinely used to screen generated
samples. Each is a binary test on a clean sample that provides neither a gradient nor a
conditioning signal to a pretrained model; a classifier over noisy states, where
required, must first be trained from its labels.

The problem is therefore conditional guidance of a frozen diffusion model with a binary label
computed on finished samples. A classifier over noisy states trained on such labels estimates the
same quantity as the estimator we use, so at the level of the estimator the method coincides with
classifier guidance \citep{dhariwal2021diffusion}. The contribution of this paper is not a new
estimator. It is a controlled finding about how the guidance is applied: with the direction, the
schedule and the mean step held fixed, the way the step magnitude is distributed across samples
changes constraint satisfaction. It is supported by the characterization of $h$ for a hard
indicator and by the distinction between data- and model-law estimands.

For a constraint set $C$, the posterior score decomposes as
\begin{equation}
\nabla_x \log p_t(x_t \mid C) = \nabla_x \log p_t(x_t) + \nabla_x \log h_t(x_t),
\qquad h_t(x) = \Pr(X_0 \in C \mid X_t = x),
\label{eq:decomp}
\end{equation}
so that conditioning a frozen model reduces to estimating a single scalar function: the
probability that a noisy state denoises into $C$.
 
A classifier over noisy states trained on such labels
\citep{dhariwal2021diffusion} estimates $h$ directly, and \citet{uehara2024understanding}
interpret classifier guidance as a Doob $h$-transform with $h$ learned by regression on
forward-noised labelled data. Other guidance methods approximate this decomposition
using information that is unavailable in the binary setting: conditioning labels during
training \citep{ho2022classifierfree}, known measurement operators
\citep{chung2023dps, song2023pigdm}, paired data for a learned generalized $h$-transform
\citep{denker2024deft}, text guidance or reward models for training-free $h$-transform
editing \citep{nguyen2025hedit}, or differentiable evaluators
\citep{bansal2023universal, yu2023freedom}. Estimators of $h$ itself differ along two
axes. Simulation-based estimators fit $h$ to endpoints of the model's own reverse
process \citep{li2024svdd, guo2026hardconstraint} or estimate its gradient by Monte Carlo
lookahead during sampling \citep{zhu2026doit}; they require base-model evaluations or
reward calls and target the model distribution rather than the data distribution.
Related gradient-guidance methods use learned return or property predictors in trajectory,
graph, and molecular diffusion \citep{janner2022diffuser, vignac2023digress, lee2023mood}.
More directly, \citet{go2026stitchvm} regress clean-sample reward-model scores onto
forward-noised latents, while the guarantees of \citet{chang2026doobmatching} require the
weight to be bounded away from zero, which excludes hard indicators.

We study this setting for hard indicator constraints, and our central observation concerns
the size of the guidance step. Several guidance methods treat it as a free parameter and rescale
the gradient to a fixed radius or norm \citep{yang2024dsg, ma2024elucidating}. The Doob
$h$-transform \citep{doob1957} does not: its drift $\nabla\log h_t=\nabla h_t/h_t$ shrinks as the
constraint signal contracts under forward noising and, at a given noise level, carries a factor
$1/h_t$ across samples. We characterize $h$ as the unique bounded solution of the backward
Kolmogorov Cauchy problem of the reverse-time diffusion among those attaining $\mathbf 1_C$
along the forward process; with the exact score and $h$, the corresponding $h$-transform yields
the constrained marginals when initialized from the conditional terminal law, and
Appendix~\ref{app:proofs} bounds the practical initialization approximation. We prove that the
variance of the constraint signal vanishes at high noise, so that little is left there for any
estimator to act on, and we learn $h$ by a regression whose training pairs are obtained by
forward-noising labelled data in closed form; the verifier labels training data and is never
called during sampling. A noisy-state classifier targets the same $h$ and, for a binary reward,
value-based decoding targets its model-law counterpart (Section~\ref{sec:estimand}), so the
analysis applies to them as well.

This magnitude matters. With the schedule and the mean step size held fixed, equalizing the
step across samples, which is what normalization does at each noise level, reduces satisfaction
on CIFAR-10 by $17.0$ to $18.5$ points across three independently trained classifiers, at
comparable constraint-conditional FID (FID against real images that satisfy the constraint),
and reweighting the normalized step by $1/\hat h$ recovers part of the loss. With a 1000-step
sampler the learned estimator has nearly identical calibration error on the model's own
trajectories and on forward-noised data, and relative to rejection sampling with the verifier,
guidance reduces the sampling cost per satisfying sample by factors of $3.0$ and $4.2$ at base
rates of $30.8\%$ and $22.5\%$, and by $53$ to $55$ at $1.7\%$.

Our contributions are as follows. (1) A controlled measurement showing that the magnitude of
the guidance step carries information: on the evaluated predicates the unnormalized drift's step
is strongly negatively correlated with $\hat h$, and with the schedule and mean step size held
fixed, equalizing it costs $17.0$ to $18.5$ points of satisfaction at comparable FID across
independently trained classifiers, which a $1/\hat h$ reweighting partially restores
(Section~\ref{sec:alloc}). (2) The theory that explains it: $h_t$ is the unique bounded
solution of the reverse-time backward Kolmogorov problem attaining $\mathbf 1_C$ along the
forward process, the variance of its signal is non-increasing in the noise level and vanishes
as $\bar\alpha_t\to0$ without spectral-gap assumptions, a noisy-state classifier estimates the
same $h$, and it differs from the target of rollout-based methods by an amount set by the score
error of the base model (Sections~\ref{sec:theory} and~\ref{sec:estimand}). (3) A guidance
method that needs only binary evaluations of clean samples: one lightweight estimator per
constraint, trained by squared-error regression without base-model evaluations, whose
log-gradient is positively aligned with a rollout reference (mean cosine $0.47$ to $0.54$ from
$t=400$; Tables~\ref{tab:field} and~\ref{tab:e1}), evaluated across pixel, latent and molecular
diffusion, including composition (Sections~\ref{sec:method} and~\ref{sec:experiments}).

\section{Background and problem setting}
\label{sec:background}

\paragraph{Diffusion models.}
The forward SDE $\mathrm{d}X_t = f(X_t,t)\,\mathrm{d}t + g(t)\,\mathrm{d}W_t$
transforms data $X_0 \sim p_0$ into noise as $t \to T$, and the reverse-time
dynamics
$\mathrm{d}X_t = [f(X_t,t) - g(t)^2 \nabla_x \log p_t(X_t)]\,\mathrm{d}t
+ g(t)\,\mathrm{d}\bar{W}_t$ use the score $\nabla_x \log p_t$, approximated by
$s_\theta(x,t)$ \citep{anderson1982reverse, song2021sde}. In the DDPM discretization
\citep{ho2020ddpm},
$\nabla_{x}\log p_t(x) \approx -\epsilon_\theta(x,t)/\sqrt{1-\bar\alpha_t}$ with
$\bar\alpha_t = \prod_{s=1}^{t}(1-\beta_s)$.
 
\paragraph{Reverse-time generator.}
For a functional $\phi$ of clean data, the posterior expectation
$u(x,t) = \mathbb{E}[\phi(X_0)\mid X_t=x]$ is governed by the generator of the
reverse-time diffusion,
\begin{equation}
\widehat{\mathcal{L}}_t u = \tfrac{1}{2}g(t)^2\Delta u
  + \big(g(t)^2\nabla_x\log p_t - f\big)\cdot\nabla_x u .
\label{eq:revgen}
\end{equation}
Unlike the forward generator, which propagates expectations of \emph{future}
states, $\widehat{\mathcal{L}}_t$ propagates expectations of the \emph{initial}
state, and depends on the data through the score.
 
\paragraph{\texorpdfstring{Doob $h$-transform.}{Doob h-transform.}}
For $h_t(x) = \Pr(X_0\in C\mid X_t=x)$, Bayes' rule gives
$p_t(x\mid C) = p_t(x)\,h_t(x)/\Pr(X_0\in C)$, so the conditional score is the
unconditional score plus $\nabla_x\log h_t$ \citep{doob1957}. Adding this drift
reweights reverse trajectories toward $C$ and is exact: started from
$p_T(\cdot\mid X_0\in C)$, the resulting process has marginals $p_t(\cdot\mid X_0\in C)$.
 
\paragraph{Problem setting.}\label{sec:setting}
 
A constraint is given only through its indicator $\mathbf{1}_C:\mathcal{X}\to\{0,1\}$,
evaluable on clean samples. The learner may evaluate $\mathbf{1}_C(x_0)$ on clean
data and noise those samples forward through the known kernel. It has no
differentiable relaxation of $\mathbf{1}_C$, no evaluation of it on noisy states $x_t$, $t>0$,
no conditioning signal in the frozen base model, and does not roll the base model
out to obtain labels. The goal is to sample from $p_0(\cdot\mid X_0\in C)$ using
the frozen score model and a learned estimate of $h$. Proofs of all results are in
Appendix~\ref{app:proofs}.
 
\section{Theory}
\label{sec:theory}

The results below rest on standard tools (time reversal, It\^o's formula, the Doob transform);
what is specific here is the discontinuous initial condition $\mathbf 1_C$, for which $h$
vanishes on part of the space, and each result makes a prediction that
Section~\ref{sec:experiments} tests.
 
\begin{proposition}[Regression target]
\label{prop:target}
For each $t$, the unique minimizer over measurable $h$ of
$\mathbb{E}\,[(h(X_t)-\mathbf{1}_C(X_0))^2]$, with $(X_0,X_t)$ drawn from the data
and the forward kernel, is $h_t(x)=\Pr(X_0\in C\mid X_t=x)$.
\end{proposition}
 
\begin{theorem}[Characterization]
\label{thm:char}
Under standard regularity conditions (Appendix~\ref{app:proofs}), for any measurable
$C$ with $0<\Pr(X_0\in C)<1$ the function $h_t(x)=\Pr(X_0\in C\mid X_t=x)$ is the
unique bounded classical solution of the Cauchy problem
\begin{equation}
\partial_t h = \widehat{\mathcal{L}}_t h, \qquad h_0 = \mathbf{1}_C,
\label{eq:cauchy}
\end{equation}
with $\widehat{\mathcal{L}}_t$ as in~\eqref{eq:revgen} and the initial condition
attained along the forward process, $h_\varepsilon(X_\varepsilon)\to\mathbf 1_C(X_0)$
as $\varepsilon\downarrow0$. Started from $p_T(\cdot\mid X_0\in C)$, the reverse-time
process with drift augmented by $g(t)^2\nabla_x\log h_t$ has marginals
$p_t(\cdot\mid X_0\in C)$ at every $t$.
\end{theorem}
 
Proposition~\ref{prop:target} is the standard characterization of the
conditional mean, and it is why binary clean labels suffice: $\mathbf{1}_C(X_0)$
is an unbiased sample of $h_t(X_t)$ at every noise level.
Theorem~\ref{thm:char} identifies the target of every guidance method in this
setting as the unique bounded solution, in the sense made precise in
Appendix~\ref{app:thm1}, of a parabolic problem whose initial condition is the
constraint indicator.
 
\begin{proposition}[Signal contraction]
\label{prop:contraction}
Let $\bar h=\Pr(X_0\in C)$ and $\sigma^2(t)=\mathrm{Var}[h_t(X_t)]$. Then
(i) $\mathbb{E}[h_t(X_t)]=\bar h$ for all $t$, and $\sigma^2(t)$ is non-increasing
in $t$; (ii) $\sigma^2(t)\le\sqrt{I(X_0;X_t)/2}$, where $I$ is mutual
information in nats; and (iii) for the variance-preserving process on
$\mathbb R^d$ with $m_2=\mathbb{E}\|X_0\|^2<\infty$,
$I(X_0;X_t)\le\tfrac{d}{2}\log\!\big(1+\tfrac{\bar\alpha_t\,m_2}{(1-\bar\alpha_t)\,d}\big)$,
so $\sigma^2(t)\to 0$ as $\bar\alpha_t\to 0$.
\end{proposition}
 
\begin{figure}[t]
\centering
\includegraphics[width=\linewidth]{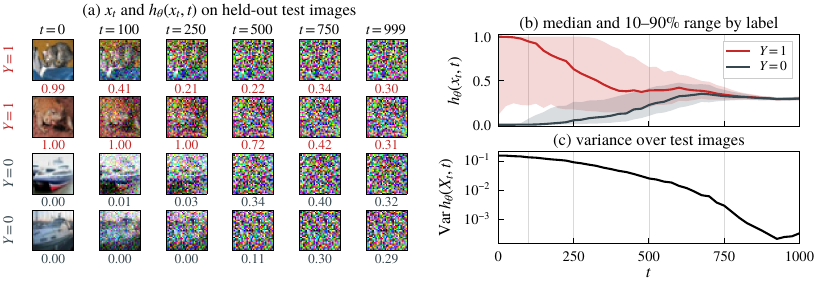}
\caption{\textbf{The learned $h_\theta$ on held-out, forward-noised CIFAR-10 images}
(red-pixel, $Y=\mathbf 1_C(x_0)$). (a) Two test images with $Y=1$ and two with $Y=0$ along
one forward path, with $h_\theta(x_t,t)$ under each. (b) Over 4,000 test images the label
groups separate at small $t$ and merge as $t$ grows. (c) The variance of the learned
$h_\theta(X_t,t)$ falls from $0.15$ at $t=0$ to $3\times10^{-4}$ at $t=900$, as
Proposition~\ref{prop:contraction} predicts for $h_t$.}
\label{fig:cifar_h}
\end{figure}

Proposition~\ref{prop:contraction} makes high-noise guidance failure a property of
the forward process rather than of any estimator: as $t$ grows, $h_t$ flattens
toward $\bar h$ and carries progressively less information about $C$. On MNIST
\citep{lecun1998mnist} digit-3 the measured $\hat\sigma^2(t)$ falls by $492\times$ between $t=100$ and
$t=900$ (Figure~\ref{fig:diagnostic}a, Appendix~\ref{app:diag}), and the mean guidance norm falls from $2.32$ to $0.225$
over the same range (Appendix~\ref{app:mnist});
Figure~\ref{fig:cifar_h} shows the same contraction for a learned $h_\theta$ on CIFAR-10.

\paragraph{Relation to the spectrum of the conditional-expectation operator.}
\citet{moreira2026spectral} study the operator $(T_tf)(x)=\mathbb{E}[f(X_0)\mid X_t=x]$ and, when
it is compact (for instance for compactly supported data), its singular system
$(\sigma_{t,k},\psi_{t,k},\phi_{t,k})$ with $\sigma_{t,1}=1$ and constant leading functions; they
show that $\sigma_{t,k}\to0$ for $k\ge2$ as $\bar\alpha_t\to0$ and locate empirically the noise
window in which guidance is effective. Since $h_t-\bar h=T_t(\mathbf 1_C-\bar h)$,
\begin{equation}
\sigma^2(t)=\sum_{k\ge2}\sigma_{t,k}^2\,\langle\mathbf 1_C,\psi_{t,k}\rangle_{p_0}^2 ,
\label{eq:spectral}
\end{equation}
so the variance of Proposition~\ref{prop:contraction} is their spectrum weighted by the
coefficients of the constraint. Proposition~\ref{prop:contraction} adds monotonicity in $t$ and an
explicit bound that requires only a finite second moment. Appendix~\ref{app:sg} compares the two
methods empirically.

\paragraph{From the regression target to the guidance field.}
Proposition~\ref{prop:target} controls $h_\theta$ in squared error, while the
sampler uses $\nabla\log h_\theta$. Under the variance-preserving kernel the ideal
field has the closed form (Appendix~\ref{app:tweedie})
\begin{equation}
\nabla_{x_t}\log h_t(x_t)=\frac{\sqrt{\bar\alpha_t}}{1-\bar\alpha_t}
\Big(\mathbb{E}[X_0\mid X_t=x_t,\,X_0\in C]-\mathbb{E}[X_0\mid X_t=x_t]\Big),
\label{eq:tweedie}
\end{equation}
which can be estimated by rolling the frozen model out from $x_t$.
Tables~\ref{tab:field} and~\ref{tab:e1} (Appendix~\ref{app:moved} and~\ref{app:field}) compare
the learned field with this reference on MNIST digit $3$: the mean cosine is $0.47$ to $0.54$
at every measured noise level from $t=400$, where most states have enough satisfying
rollouts to estimate the reference (Appendix~\ref{app:field}), against a chance level near zero. The logarithm's
sensitivity to small $h$ has little effect in practice: sweeping $\delta$ from $10^{-10}$ to
$10^{-2}$ leaves satisfaction at $94.8\pm0.5\%$, and the fraction of held-out states with
$h_\theta<0.01$ falls from 89\% at $t=50$ to 0\% from $t=700$ (Appendix~\ref{app:mnist}).

\section{Method}
\label{sec:method}
 
\paragraph{Objective.}
We parameterize $h_\theta(x,t)\in(0,1)$ by a small network with a sigmoid output
and minimize
$\mathcal{L}=\mathcal{L}_{\mathrm{bdry}}+\mathcal{L}_{\mathrm{sup}}
+\lambda_{\mathrm{fc}}\mathcal{L}_{\mathrm{fc}}$ with $\lambda_{\mathrm{fc}}=0.1$:
\begin{align}
\mathcal{L}_{\mathrm{sup}} &= \mathbb{E}_{x_0,t,\epsilon}\big[(h_\theta(x_t,t)-\mathbf{1}_C(x_0))^2\big],
\qquad
\mathcal{L}_{\mathrm{bdry}} = \mathbb{E}_{x_0}\big[(h_\theta(x_0,0)-\mathbf{1}_C(x_0))^2\big],\\
\mathcal{L}_{\mathrm{fc}} &= \mathbb{E}_{x_0,t}\big[(h_\theta(x_t,t)-\mathrm{sg}[h_\theta(x'_{t+1},t+1)])^2\big],
\end{align}
where $x_t$ and $x'_{t+1}$ are independent forward noisings of the same $x_0$
and $\mathrm{sg}$ is stop-gradient. $\mathcal{L}_{\mathrm{sup}}$ alone identifies
$h$ (Proposition~\ref{prop:target}) and suffices in practice (Appendix~\ref{app:ablation});
it is the core of the method. The other two are optional regularizers, used for every head
reported in Section~\ref{sec:experiments}:
$\mathcal{L}_{\mathrm{bdry}}$ places weight on the discontinuous initial
condition, which uniform timestep sampling rarely visits, and
$\mathcal{L}_{\mathrm{fc}}$ smooths the estimate across noise levels
(Appendix~\ref{app:objective}). Every training pair is produced in closed form from
labelled data, so training requires no base-model evaluations. Learning the
guidance field directly would amount to training a score model on the
constraint-satisfying samples alone, a regression in the data dimension from the
positive fraction of the data (6.1\% on Stable Diffusion); the scalar head is
trained on every labelled sample, positive and negative, and its gradient is a
conservative field by construction.
 
\paragraph{Sampling.}
In the DDPM parameterization we replace the noise prediction by
\begin{equation}
\tilde\epsilon(x_t,t)=\epsilon_\theta(x_t,t)-\gamma\sqrt{1-\bar\alpha_t}\,
\nabla_{x_t}\log\big(h_\theta(x_t,t)+\delta\big),
\label{eq:sampler}
\end{equation}
with guidance scale $\gamma$ and $\delta=10^{-8}$ unless stated. At $\gamma=1$ and $\delta=0$, with an
exact score and an exact $h$, this is the $h$-transform of Theorem~\ref{thm:char}
(Appendix~\ref{app:sampler}); as is standard for guidance scales \citep{dhariwal2021diffusion},
we use $\gamma>1$, and $\delta$ keeps $\log h$ finite where $h\approx 0$. Each step
costs one forward and one backward pass of $h_\theta$; there is no candidate set
and no rollout. Algorithm~\ref{alg:main} (Appendix~\ref{app:moved}) gives both procedures.

\paragraph{Composition.}
For $C_1\cap C_2$ with separately trained heads, we guide with
$w_1\nabla\log h^{(1)}+w_2\nabla\log h^{(2)}$. With $w_1=w_2=1$ this is the exact
joint correction when $\mathbf{1}_{C_1}(X_0)$ and $\mathbf{1}_{C_2}(X_0)$ are
conditionally independent given $X_t$ at every $t$ (Appendix~\ref{app:sampler});
otherwise the weights trade off the two constraints, and the composed drift is not guaranteed
to reach the joint constraint; Appendix~\ref{app:depcomp} tests a strongly dependent pair.
 
\section{Data and model estimands of \texorpdfstring{$h$}{h}}
\label{sec:estimand}
 
With a binary terminal reward, $h$ is the value $\mathbb{E}[\mathbf 1_C(X_0)\mid X_t]$ targeted by
value-learning methods \citep{li2024svdd, yang2021fudge, lawson2022sixo,
uehara2024understanding}, and our
$\mathcal{L}_{\mathrm{sup}}$ shares its gradient in expectation with consistency
value regression. The difference lies in which conditional law the expectation
is taken under. Rollout-based estimators \citep{li2024svdd, guo2026hardconstraint}
regress onto endpoints of the \emph{model's} reverse process started at $x$, and
so estimate
$h^{\mathrm{model}}_t(x)=\Pr_{\mathrm{model}}(X_0\in C\mid X_t=x)$. Forward
noising of labelled data, as here and in \citet{chang2026doobmatching}, estimates
$h^{\mathrm{data}}_t=h_t$.
 
\begin{proposition}[Estimand gap]
\label{prop:estimand}
Under the assumptions of Appendix~\ref{app:prop3}, $h^{\mathrm{model}}$
solves~\eqref{eq:cauchy} with $\nabla_x\log p_t$ replaced by
the model score $s_\theta$ in $\widehat{\mathcal{L}}_t$. Consequently
$h^{\mathrm{model}}=h^{\mathrm{data}}$ for every $C$ if the model's reverse
process reproduces the data posterior, and in general the two differ by
an amount governed by the score error $s_\theta-\nabla_x\log p_t$ rather than by
either estimator.
\end{proposition}
 
With an exact score, guiding with $h^{\mathrm{data}}$ samples the data distribution
restricted to $C$, the reference of constraint-conditional FID. With the model's
score only $h^{\mathrm{model}}$ gives an exact $h$-transform of the process being
sampled; Section~\ref{sec:gap} compares the estimator's calibration under the two laws for a real model. The two targets also differ in cost:
forward noising gives every labelled sample an independent training pair at every
noise level with no base-model evaluations, whereas the SVDD-MC targets of
Section~\ref{sec:cost} consumed $5{,}000{,}000$ base-model forward passes for
$5{,}000$ independent endpoint outcomes. Equal gradients in expectation do not
imply equal estimator variance at finite sample size.
 
Estimators of $h$ therefore fall into a two-by-two grid: rollout targets with soft
rewards \citep{li2024svdd}, rollout targets with hard constraints
\citep{guo2026hardconstraint}, forward-noising targets with weights bounded below
\citep{chang2026doobmatching}, and forward-noising targets for hard indicators, the cell of
noisy-state classifier guidance, whose target this work characterizes and studies where $h$
approaches zero (Table~\ref{tab:grid}, Appendix~\ref{app:moved}). The analysis concerns $h^{\mathrm{data}}$
itself, so it applies to any estimator of it, including a classifier trained by cross-entropy
on the same labels, whose loss has the same minimizer: $\nabla\log h=\nabla h/h$ shrinks at high noise (by~\eqref{eq:tweedie},
$\|\nabla\log h_t\|\le\sqrt{\bar\alpha_t}\,D/(1-\bar\alpha_t)$ for data of diameter
$D$) and carries a factor $1/h$ at each noise level, whereas normalized guidance applies a
step of fixed magnitude to every sample at every step. Section~\ref{sec:alloc} measures both
effects.

\section{Experiments}
\label{sec:experiments}
 
We evaluate in three settings where binary constraints arise: image predicates that no
pretrained model conditions on, where we compare against a matched classifier and against
rejection sampling and measure how the drift allocates the step
(Sections~\ref{sec:cifar}--\ref{sec:cost}); property filters in molecular design, including composition of
independently trained constraints (Section~\ref{sec:mol}); and text-to-image generation with
Stable Diffusion~1.5 (Section~\ref{sec:sd}).

\paragraph{Protocol.}
The base model is frozen throughout. Satisfaction is the fraction of samples that
pass the predicate used for labelling (for the MNIST digits and the CIFAR-10 classes, a
separately trained classifier judges the samples), with the standard deviation over sampling seeds for a single trained head
(Wilson 95\% intervals \citep{wilson1927} where stated); FID \citep{heusel2017fid}
uses real constraint-satisfying images (constraint-conditional) and a general
reference set (full-distribution).
Classifier-guidance baselines use a classifier over noisy states trained on the
identical labels, optimizer and budget, at matched capacity, and guide with the
standard rule $\epsilon\leftarrow\epsilon-\gamma\sqrt{1-\bar\alpha_t}\,g$,
$g=\nabla_{x_t}\log p_\phi(y{=}1\mid x_t)$; the normalized variant divides $g$ by $\|g\|$ per
sample. DPS requires a differentiable measurement operator, which a binary evaluation does not
provide. Appendix~\ref{app:details} gives every setting.
 
\subsection{Matched-capacity comparison with classifier guidance}
\label{sec:cifar}
 
We define four predicates on CIFAR-10 \citep{krizhevsky2009cifar} images as
deterministic functions of pixel statistics, with positive rates of 24.5--26.8\% on
the training set: red-pixel fraction, Sobel edge density, upper-half luminance,
and low saturation (Appendix~\ref{app:cifardetails}). None exists in the
conditioning space of the base model, the DDPM of \citet{ho2020ddpm}.
Our $h_\theta$ has 675{,}969 parameters and the classifier 676{,}098, and a guided
sample costs 0.404\,s with either.
\begin{figure}[t]
\centering
\includegraphics[width=\linewidth]{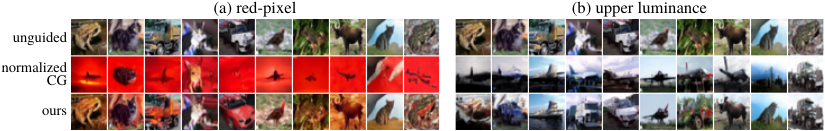}
\caption{\textbf{Paired CIFAR-10 samples at $\gamma=5$.} Each column starts from the same
$x_T$ and uses the same sampling noise in every row. The figure shows the first ten samples
generated without selection; across the full 64-sample batch, every guided sample in both rows
satisfies the constraint. The middle row, the matched classifier under the normalized rule,
achieves this by collapsing to a few image types: a red field with a small object in (a),
dark-ground aircraft scenes in (b). Guidance with $h_\theta$ keeps the content of the unguided
sample in (a) and produces varied scenes in (b).}
\label{fig:cifar_samples}
\end{figure}
Table~\ref{tab:cifar} compares our head with the matched classifier under both rules. With
the standard unnormalized drift, the two estimators reach at least $95.9\%$ satisfaction from $\gamma=5$ and
constraint-conditional FID within $6.8$ of each other, as Section~\ref{sec:estimand} predicts
for two estimates of the same $h$. Normalizing the classifier's drift keeps satisfaction at
100\% from $\gamma=5$ but gives $1.6$ to $3.2\times$ the constraint-conditional FID of our
drift at $\gamma=5$ (paired samples in Figure~\ref{fig:cifar_samples}); unlike the unnormalized drift,
whose magnitude decays as the signal contracts, the normalized step is largest at $t=999$
(Section~\ref{sec:alloc}). Appendix~\ref{app:normfront} extends the normalized rule to
$\gamma\in[0.1,1]$, so that the rules can be compared at equal satisfaction and not only at equal
$\gamma$.

\begin{table}[t]
\caption{Matched classifier under the standard and normalized rules, and ours, on four
CIFAR-10 predicates (matched in capacity, labels, optimizer, budget and inference cost).
Satisfaction (\%) and FID against constraint-satisfying test images (FID$_{\mathrm{c}}$);
FID$_{\mathrm{f}}$ in Tables~\ref{tab:cifarfull} and~\ref{tab:cgstd}. Mean over 3 seeds $\times$ 1{,}000
samples. Unguided satisfaction in parentheses.}
\label{tab:cifar}
\begin{center}
\small
\begin{tabular}{llrrrrrr}
\toprule
 & & \multicolumn{3}{c}{Satisfaction} & \multicolumn{3}{c}{FID$_{\mathrm{c}}$}\\
Predicate & $\gamma$ & CG & CG-norm & Ours & CG & CG-norm & Ours \\
\midrule
Red-pixel (31.3)& 5 & 100.0 & 100.0 & 100.0 & 62.8 & 187.0 & 58.2\\
 & 15 & 100.0 & 100.0 & 100.0 & 98.2 & 212.3 & 93.1\\
Upper luminance (23.1)& 5 & 100.0 & 100.0 & 100.0 & 43.4 & 81.4 & 43.9\\
 & 15 & 100.0 & 100.0 & 100.0 & 50.1 & 87.7 & 52.7\\
Edge density (16.5)& 5 & 99.7 & 100.0 & 100.0 & 52.5 & 96.9 & 59.3\\
 & 15 & 100.0 & 100.0 & 100.0 & 72.7 & 109.5 & 78.8\\
Low saturation (16.8)& 5 & 99.5 & 100.0 & 95.9 & 48.6 & 101.6 & 45.9\\
 & 15 & 100.0 & 100.0 & 99.9 & 71.2 & 116.8 & 73.2\\
\bottomrule
\end{tabular}
\end{center}
\end{table}
 
\subsection{How the unnormalized drift allocates the step}
\label{sec:alloc}

By~\eqref{eq:tweedie} and $\nabla\log h=\nabla h/h$, the unnormalized log-probability drift
has two properties that a normalized step removes: its size varies over the noise schedule, and
at each step it varies across samples, with a factor $1/h$. We measure both using the matched
classifier's estimate of this drift, so that the result concerns the sampling rule rather than
our estimator, on
red-pixel and upper luminance at $\gamma=2$ (Appendix~\ref{app:alloc}). Over the schedule, the mean step per
unit $\gamma$, $\bar m(t)$, the mean of $\sqrt{1-\bar\alpha_t}\,\|g\|$ over samples, peaks near
$t=500$ on red-pixel and $t=700$ on upper luminance and is $4.4$ and $3.7\times$ smaller at
$t=999$, whereas the normalized step $\sqrt{1-\bar\alpha_t}$ is largest at $t=999$. Across
samples, step size and $\hat h(x_t)$ have Spearman correlation between $-0.60$ and
$-0.998$ at $t\in\{799,599,399,199\}$ on both predicates (250 trajectories; $p<10^{-24}$;
Table~\ref{tab:spearman}).

To isolate the allocation across samples, we keep the normalized direction and give every
sample the step $\bar m(t)$ measured along the standard-rule trajectories, so that the
schedule and the mean step are those of the standard rule (Table~\ref{tab:alloc}).
Satisfaction falls by $18.4$, $18.5$ and $17.7$ points on red-pixel and by $17.2$, $17.3$ and
$17.0$ on upper luminance for three independently trained classifiers, with FID$_{\mathrm{c}}$ within $1.2$ of the standard rule in every
case: the loss is a property of the rule rather than of one classifier. Weighting each sample's step by
$(1/\hat h_i)/\mathrm{mean}_j(1/\hat h_j)$, which keeps the mean step, recovers
$35$ to $46\%$ and $60$ to $69\%$ of the loss, again at comparable FID$_{\mathrm{c}}$; the rest is carried by the
variation of $\|\nabla\hat h\|$ across samples.

Appendix~\ref{app:alloc2} repeats the test in $26$ further configurations
(Table~\ref{tab:allocconfigs}): at $\gamma=1$ and $5$, on edge density and low saturation, with
$100$- and $50$-step samplers, with our head $h_\theta$ in place of a classifier, on the class
constraints cat and ship, with the estimator of Spectral Guidance, on a conjunction of two
negatively dependent predicates, and on the molecular filter of Section~\ref{sec:mol}. The equal
step lowers mean satisfaction in every one, by $1.1$ to $22.4$ points (Tables~\ref{tab:alloc2},
\ref{tab:allocclass}, \ref{tab:sgalloc}, \ref{tab:depalloc} and~\ref{tab:qm9alloc}); its size
varies with the constraint.

\begin{table}[t]
\caption{The step at a fixed schedule and mean step size, CIFAR-10, $\gamma=2$, for independently
trained classifiers (same recipe; training seeds $0$, $1$ and $2$). All rules use the classifier's
gradient direction; ``equal'' and ``$1/\hat h$'' give each sample a step whose mean over samples is
that classifier's standard-rule $\bar m(t)$. Classifier 1 is the matched classifier of Section~\ref{sec:cifar}: mean over sampling seeds $0$ and $1$
($1/\hat h$ upper luminance: seed $0$); classifiers 2 and 3: sampling seed $0$; $1{,}000$ samples
per run.}
\label{tab:alloc}
\begin{center}
\small
\begin{tabular}{llrrrrrr}
\toprule
 & & \multicolumn{3}{c}{Satisfaction (\%)} & \multicolumn{3}{c}{FID$_{\mathrm{c}}$}\\
Predicate & Classifier & Standard & Equal & $1/\hat h$ & Standard & Equal & $1/\hat h$\\
\midrule
Red-pixel & 1 & 99.0 & 80.6 & 89.0 & 52.60 & 51.85 & 53.06\\
 & 2 & 98.2 & 79.7 & 86.2 & 53.25 & 53.04 & 54.11\\
 & 3 & 98.5 & 80.8 & 88.0 & 53.21 & 53.34 & 53.75\\
\midrule
Upper luminance & 1 & 99.7 & 82.5 & 94.3 & 39.61 & 40.73 & 39.34\\
 & 2 & 98.5 & 81.2 & 91.6 & 39.49 & 40.29 & 39.76\\
 & 3 & 99.7 & 82.7 & 94.5 & 39.55 & 40.32 & 40.13\\
\bottomrule
\end{tabular}
\end{center}
\end{table}

\subsection{Calibration under the data and model laws}
\label{sec:gap}

We compare the calibration of the learned head under the two laws of
Proposition~\ref{prop:estimand} for a real model. Along $2{,}000$ unguided trajectories we record
$h_\theta(x_t,t)$ at ten noise levels and score it against each trajectory's outcome
$\mathbf 1_C(x_0)$, which measures its calibration to $h^{\mathrm{model}}$; on forward-noised
test images at the same $t$, the same score measures calibration to $h^{\mathrm{data}}$.
With the 1000-step sampler the head has nearly identical calibration error under the two laws:
mean expected calibration error $0.051$ and $0.051$ on red-pixel, and $0.021$ and $0.024$ on
upper luminance.
Sampling with 100 steps moves the model's red-pixel rate from $31.5\%$ to $46.0\%$ and
raises the model-law error to $0.155$, while the data-law error is $0.049$; upper
luminance, whose rate stays at $21.8\%$, keeps $0.023$ against $0.024$. The difference in
calibration follows the shift of the model's law on the constraint, and it does not prevent
guidance: with 100
steps, $\gamma=2$ reaches $99.5\%$ on both predicates (Appendix~\ref{app:alloc}).

\subsection{Guidance and rejection sampling}
\label{sec:reject}

Rejection sampling with the verifier draws exactly from the model's constrained law, at
$1/p$ samples per accepted one for a pass rate $p$. On red-pixel and upper luminance
($p=30.8\%$ and $22.5\%$ over $4{,}000$ and $4{,}500$ unguided samples), the first
$1{,}000$ accepted samples have FID$_{\mathrm{c}}$ $48.4$ and $42.4$ (Appendix~\ref{app:alloc}); our guidance at
$\gamma=2$ reaches $97.2\%$ and $99.5\%$ at FID$_{\mathrm{c}}$ $48.8$ and $39.2$
(Table~\ref{tab:cifarfull}), and since a guided step costs about $6\%$ more than an
unguided one on the same GPU, it needs $3.0$ and $4.2\times$ less sampling time per
satisfying sample. For a rare constraint we raise the red-pixel threshold to the 99th
percentile of the training statistic ($1.00\%$ positive; Appendix~\ref{app:alloc}). The
model passes it $1.7\%$ of the time (Wilson interval $1.2$--$2.4\%$, $n=2{,}000$), so
rejection draws about $59$ samples per accepted one; guidance at $\gamma=2$ reaches
$95.5$ to $99.0\%$ over two training seeds of the head (the matched classifier: $95.4$ to
$96.0\%$), $53$ to $55\times$ less time per satisfying sample ($38$ to $77\times$ over the
Wilson interval). Bounds on the guidance error grow with the inverse probability of the
condition \citep{sahu2026reliable}; at $1\%$ both estimators still exceed $95\%$.

\subsection{Supervision and inference cost}
\label{sec:cost}
 
Against SVDD \citep{li2024svdd} on MNIST digit-3, with value networks matched at
122{,}657 parameters, our training targets need no base-model evaluations, where the
rollout targets of SVDD-MC consume 5{,}000{,}000 base-model forward passes (340\,s against
625\,s of training). Our sampler runs at 0.050\,s/sample with 95.1\% satisfaction, against
0.294 and 1.048\,s/sample for SVDD-PM with $M=5$ and $M=20$ candidates drawn in proportion
to value (92.5\% and 95.2\%); keeping the best candidate reaches 100.0\% at $5.7\times$ our
cost (Appendix~\ref{app:svdd}).
 
\subsection{Molecular property constraints}
\label{sec:mol}
 
On QM9 \citep{ramakrishnan2014qm9} with the GDSS base model \citep{jo2022gdss},
we use RDKit\footnote{\url{https://www.rdkit.org}} predicates on decoded molecules: ring presence, and a
Lipinski-type filter \citep{lipinski1997} (molecular weight below 125, at most one
hydrogen-bond donor, at most two acceptors, thresholds scaled to QM9's size range
and giving a 35.3\% positive rate). Guidance is added to the atom-feature score of
GDSS; validity is at least 99.9\% in every run (Appendix~\ref{app:qm9}). The
unguided baseline and all guided runs use the same Euler--Maruyama discretization of the
GDSS reverse SDE (Appendix~\ref{app:qm9details}).
Table~\ref{tab:mol} supports three observations. First, on the Lipinski-type filter
our method raises satisfaction from 54.3\% to 96.8\% at $\gamma=10$ while retaining
83.2\% uniqueness, with a satisfaction--diversity trade-off controlled by
$\gamma$. Second, SVDD-PM reaches 96.0\% at 7.4$\times$ the cost of GDSS's predictor--corrector
sampler measured on the same GPU (8.24 against 1.12\,s/mol), since every step
decodes and scores $M=5$ candidates, whereas our sampler adds one evaluation of the head
and its gradient per step. Third, composition works without joint training:
guiding with two independently trained heads raises the joint rate from 39.6\% to
78.3\%, and at $\gamma=20$ it matches the product of the guided marginals
(96.3\% $\times$ 81.6\% $=$ 78.6\%; Table~\ref{tab:qm9comp}), consistent with the two
properties being nearly independent under the model (unguided: 39.6\% joint, 41.6\% product).

\begin{table}[t]
\caption{QM9, 3 seeds $\times$ 1{,}000 molecules. Satisfaction, uniqueness and
novelty (\%). The composed rows guide with $\log h_{\mathrm{Lip}}+\log h_{\mathrm{ring}}$
from separately trained heads; their satisfaction is the joint rate. All rows use the same
Euler--Maruyama sampler. SVDD-PM: 1 seed $\times$ 500 molecules (Appendix~\ref{app:qm9details}).}
\label{tab:mol}
\begin{center}
\begin{tabular}{llrrrr}
\toprule
Constraint & Method & $\gamma$ / $M$ & Sat. & Uniq. & Novel\\
\midrule
Lipinski-type & Unguided & 0 & 54.3 & 98.0 & 98.3\\
 & Ours & 2 & 88.7 & 92.4 & 96.2\\
 & Ours & 10 & 96.8 & 83.2 & 94.5\\
 & Ours & 50 & 99.5 & 63.2 & 92.8\\
 & SVDD-PM & $M=5$ & 96.0 & 97.8 & 97.1\\
\midrule
Lipinski $\wedge$ ring & Unguided & 0 & 39.6 & 98.0 & 98.3\\
 & Composed & 5 & 72.4 & 92.3 & 96.0\\
 & Composed & 20 & 78.3 & 90.8 & 95.9\\
\bottomrule
\end{tabular}
\end{center}
\end{table}
 
\subsection{Scale: Stable Diffusion 1.5}
\label{sec:sd}
 
On Stable Diffusion~1.5 \citep{rombach2022ldm} with 50-step DDIM
\citep{song2021ddim} and classifier-free guidance scale 7.5, we train $h_\theta$ on a fixed
30{,}000-image COCO \citep{lin2014coco} subset labelled by a grayscale predicate (6.1\% positive) and guide in
latent space. Guidance scales are larger than in pixel space because
classifier-free guidance inflates the noise prediction.
Satisfaction rises from 7.2\% to 60.0\%, monotonically in $\gamma$ up to $\gamma=75$
(Tables~\ref{tab:sd} and~\ref{tab:sdfull}; samples in Figure~\ref{fig:sd_samples}, appendix), with
constraint-conditional FID improving from 109.9 to 98.2. Derivative-free
candidate selection reaches 9.0\% at 16$\times$ our sampling cost. The normalized
rule is more sensitive to the scale: swept from $\gamma=1$ to $75$, the matched classifier rises
from 12.3\% at $\gamma=1$ to 65.1\% at $\gamma=10$ and declines to 24.7\% at $\gamma=75$, while our
drift rises monotonically over its tested range (Table~\ref{tab:sdfull}, Appendix~\ref{app:sd}). The two rules are tuned separately
here and differ in the size of the step and in its profile over time, so this comparison does not
isolate how the step is allocated across samples; the matched-budget result of
Section~\ref{sec:alloc} is established on CIFAR-10 and QM9, and we do not claim it for latent
diffusion. On a separate set of
200 prompts (one seed, 500 images; Table~\ref{tab:sdstd}), the classifier under the standard rule reaches $28.8$,
$46.6$ and $54.4\%$ at $\gamma=10$, $35$ and $75$, against $30.0$, $48.8$ and $58.0\%$ for our
head, while the normalized rule falls from $61.8\%$ at $\gamma=10$ to $25.4\%$ at $\gamma=75$.

\section{Related work}
\label{sec:related}
 
\paragraph{Guidance with gradients or labels.}
Classifier guidance \citep{dhariwal2021diffusion} and classifier-free guidance
\citep{ho2022classifierfree} require noisy-state classifiers or conditional
training; DPS and $\Pi$GDM \citep{chung2023dps, song2023pigdm} require
a known measurement operator, and Universal Guidance and FreeDoM
\citep{bansal2023universal, yu2023freedom} differentiable evaluators. The decomposition~\eqref{eq:decomp} underlies
conditional sampling in \citet{song2021sde}, and \citet{didi2023framework} cast
conditional generation, including motif scaffolding, as a Doob $h$-transform; DEFT
\citep{denker2024deft} learns a generalized $h$-transform from paired data, h-Edit
\citep{nguyen2025hedit} applies the $h$-transform to editing, and ensemble Kalman
guidance \citep{zheng2025enkg} is derivative-free for black-box inverse problems
but propagates an ensemble of particles.
Self-attention guidance \citep{hong2023sag} uses the denoiser's self-attention maps to
select the regions it blurs and guides against, and perturbed-attention guidance
\citep{ahn2024pag} perturbs the self-attention maps themselves; both improve sample quality and
take no constraint as input, and CFG++ \citep{chung2025cfgpp} modifies classifier-free guidance for a
conditionally trained model; they address sample quality or conditional models and can be combined
with a constraint drift.
 
\paragraph{Analyses of guidance.} Guidance does not in general sample the intended tilted
distribution \citep{chidambaram2024guidance}, and restricting it to a noise interval
improves quality \citep{kynkaanniemi2024interval}; training-free methods often rescale the
guidance gradient, for instance to a fixed radius \citep{yang2024dsg, ma2024elucidating}, the choice whose cost
Section~\ref{sec:alloc} measures. Spectral Guidance \citep{moreira2026spectral} learns the leading
singular functions of the conditional-expectation operator from independently noised views and
projects any clean-data signal onto them; its released sampler rescales each sample's
log-gradient to a norm in $[0.1,0.3]$. Section~\ref{sec:theory} relates its spectral analysis to
Proposition~\ref{prop:contraction} and Appendix~\ref{app:sg} compares with it. Recent theory characterizes when linear score combinations
realize projective composition and when they fail \citep{bradley2025projective}.

\paragraph{Value functions and twisted sampling.}
SVDD \citep{li2024svdd} guides diffusion by soft value functions with candidate
selection at each step; FUDGE \citep{yang2021fudge}, SIXO \citep{lawson2022sixo},
controlled decoding \citep{mudgal2024controlled}, twisted SMC, including for indicator
targets \citep{zhao2024probabilistic}, and its amortized form in discrete diffusion
\citep{kim2026cdm} learn values or twists from base-model samples or SMC approximations of
the target, and Value Matching
\citep{jensen2026value} learns a value for reward tilting from on-policy rollouts. These
methods differ in how the value or twist is estimated and in whether inference uses candidate
selection, particles, or token-level reweighting.
Reward fine-tuning instead updates the generator: Flow-GRPO \citep{liu2025flowgrpo} applies
online policy-gradient reinforcement learning to flow models, DiffusionNFT
\citep{zheng2026diffusionnft} fine-tunes on the forward process from the model's own online samples,
and ReFL \citep{xu2023imagereward} backpropagates a differentiable reward into the generator. Noise
Hypernetworks \citep{eyring2025hypernoise} keep a distilled generator frozen and train a network
that modifies its initial noise, by maximizing a differentiable reward through the generator.
These methods adapt the generator or its input using samples drawn from it, and are an alternative
route to the same goal; we keep the generator and its input unchanged and train one small head per
constraint from labelled data, with a check that need not be differentiable. We do not compare
with them empirically. With an exact score and $h$, the constrained score of
Theorem~\ref{thm:char} is a candidate teacher score for distribution-matching distillation
\citep{yin2024dmd}, which we do not evaluate.
 
\paragraph{\texorpdfstring{Learning $h$ directly.}{Learning h directly.}}
\citet{guo2026hardconstraint} learn $h$ for hard constraints with a martingale loss
on trajectories of the pretrained model and give total-variation and Wasserstein
guarantees; \citet{chang2026doobmatching} learn $h$ by gradient-penalized
regression on forward-noised samples with Wasserstein guarantees for weights
bounded below; \citet{go2026stitchvm} regress neural reward-model scores on
forward-noised latents and guide with the value gradient;
\citet{uehara2024understanding} also describe guidance with the gradient of a value learned
by Monte Carlo regression on samples of the pretrained model; and \citet{zhu2026doit}
estimate the Doob correction for an event by Monte Carlo lookahead without
training. Section~\ref{sec:estimand} and Table~\ref{tab:grid} position this work.
 
\section{Discussion and conclusion}

Guidance from binary labels is a Doob $h$-transform, and its effectiveness depends less on
the estimator of $h$ than on the drift constructed from it: $\nabla h/h$ acts at noise levels
where the signal has not contracted, its step is negatively correlated with $\hat h$ across
samples, and at a matched budget this allocation accounts for $17.0$ to $18.5$ points of
satisfaction on CIFAR-10 that equal per-sample steps forgo, for each of the three classifiers
we trained. In the regimes we evaluate, guidance is substantially cheaper per
satisfying sample than rejection sampling, with the advantage increasing for the rare constraint
we study. One estimator per constraint, trained once from binary labels and combined by adding
log-gradients, guides a frozen model at a modest overhead over unguided sampling ($5$ to $6\%$ on
CIFAR-10, $27\%$ on Stable Diffusion) without invoking the verifier during sampling. In our
experiments, rescaling the guidance gradient to a fixed size discards the information carried by
the step size, and reweighting a normalized step by $1/\hat h$ restores part of it. Natural
extensions are costly verifiers, such as docking, and rarer constraints.

\subsection*{AI use statement}

Generative AI tools were used to aid and polish the writing and to help draft sections of the
paper from the authors' notes, results and instructions. The research idea and the method
originate with the authors, who ran all experiments and produced all reported results. All
AI-assisted text was reviewed and edited by the authors. Reported numerical results were
checked against the corresponding experimental records, and references and citation-supported
claims were manually reviewed against cited public records. The authors take full responsibility
for the content of this work, including all text, claims, and results.

\subsection*{Ethics statement}

This work uses public datasets (MNIST, CIFAR-10, COCO and QM9) and publicly released
pretrained models, and involves no human subjects. The method steers a frozen generator
toward outputs that pass a user-supplied check. Like other guidance methods it could be
pointed at checks that encode harmful properties, and it inherits the biases of the base
model and of the labelled data used to train each head.

\subsection*{Reproducibility statement}

Appendix~\ref{app:proofs} proves every result of Sections~\ref{sec:theory}
and~\ref{sec:estimand} under the assumption stated there. Appendix~\ref{app:details}
specifies each base model, head architecture, optimizer, training budget, sampler,
predicate with its thresholds, FID reference set and timing protocol, and
Appendix~\ref{app:results} reports the full sweeps behind every table of the main text,
with seeds and sample counts in each caption. Code and trained heads will be released upon publication.

\bibliography{references}

\begin{thebibliography}{60}
\providecommand{\natexlab}[1]{#1}
\providecommand{\url}[1]{\texttt{#1}}
\expandafter\ifx\csname urlstyle\endcsname\relax
  \providecommand{\doi}[1]{doi: #1}\else
  \providecommand{\doi}{doi: \begingroup \urlstyle{rm}\Url}\fi

\bibitem[Ahn et~al.(2024)Ahn, Cho, Min, Kim, Jang, Kim, Park, Jin, and
  Kim]{ahn2024pag}
Donghoon Ahn, Hyoungwon Cho, Jaewon Min, Jungwoo Kim, Wooseok Jang, SeonHwa
  Kim, Hyun~Hee Park, Kyong~Hwan Jin, and Seungryong Kim.
\newblock Self-rectifying diffusion sampling with perturbed-attention guidance.
\newblock In \emph{European Conference on Computer Vision}, 2024.

\bibitem[Anderson(1982)]{anderson1982reverse}
Brian D.~O. Anderson.
\newblock Reverse-time diffusion equation models.
\newblock \emph{Stochastic Processes and their Applications}, 12\penalty0
  (3):\penalty0 313--326, 1982.

\bibitem[Bansal et~al.(2024)Bansal, Chu, Schwarzschild, Sengupta, Goldblum,
  Geiping, and Goldstein]{bansal2023universal}
Arpit Bansal, Hong-Min Chu, Avi Schwarzschild, Roni Sengupta, Micah Goldblum,
  Jonas Geiping, and Tom Goldstein.
\newblock Universal guidance for diffusion models.
\newblock In \emph{International Conference on Learning Representations}, 2024.

\bibitem[Bradley et~al.(2025)Bradley, Nakkiran, Berthelot, Thornton, and
  Susskind]{bradley2025projective}
Arwen Bradley, Preetum Nakkiran, David Berthelot, James Thornton, and Joshua~M.
  Susskind.
\newblock Mechanisms of projective composition of diffusion models.
\newblock In \emph{Proceedings of the 42nd International Conference on Machine
  Learning}, volume 267 of \emph{Proceedings of Machine Learning Research},
  pp.\  5319--5354, 2025.

\bibitem[Chang et~al.(2026)Chang, Duan, Jiao, Li, Xu, and
  Yang]{chang2026doobmatching}
Jinyuan Chang, Chenguang Duan, Yuling Jiao, Ruoxuan Li, Yi~Xu, and
  Jerry~Zhijian Yang.
\newblock Inference-time alignment for diffusion models via variationally
  stable {Doob's} matching.
\newblock \emph{arXiv preprint arXiv:2601.06514}, 2026.

\bibitem[Chidambaram et~al.(2024)Chidambaram, Gatmiry, Chen, Lee, and
  Lu]{chidambaram2024guidance}
Muthu Chidambaram, Khashayar Gatmiry, Sitan Chen, Holden Lee, and Jianfeng Lu.
\newblock What does guidance do? {A} fine-grained analysis in a simple setting.
\newblock In \emph{Advances in Neural Information Processing Systems},
  volume~37, pp.\  84968--85005, 2024.

\bibitem[Chung et~al.(2023)Chung, Kim, McCann, Klasky, and Ye]{chung2023dps}
Hyungjin Chung, Jeongsol Kim, Michael~T. McCann, Marc~L. Klasky, and Jong~Chul
  Ye.
\newblock Diffusion posterior sampling for general noisy inverse problems.
\newblock In \emph{International Conference on Learning Representations}, 2023.

\bibitem[Chung et~al.(2025)Chung, Kim, Park, Nam, and Ye]{chung2025cfgpp}
Hyungjin Chung, Jeongsol Kim, Geon~Yeong Park, Hyelin Nam, and Jong~Chul Ye.
\newblock {CFG++}: Manifold-constrained classifier free guidance for diffusion
  models.
\newblock In \emph{International Conference on Learning Representations}, 2025.

\bibitem[Cover \& Thomas(2006)Cover and Thomas]{cover2006elements}
Thomas~M. Cover and Joy~A. Thomas.
\newblock \emph{Elements of Information Theory}.
\newblock John Wiley \& Sons, second edition, 2006.

\bibitem[Denker et~al.(2024)Denker, Vargas, Padhy, Didi, Mathis, Dutordoir,
  Barbano, Mathieu, Komorowska, and Lio]{denker2024deft}
Alexander Denker, Francisco Vargas, Shreyas Padhy, Kieran Didi, Simon Mathis,
  Vincent Dutordoir, Riccardo Barbano, Emile Mathieu, Urszula~Julia Komorowska,
  and Pietro Lio.
\newblock {DEFT}: Efficient fine-tuning of diffusion models by learning the
  generalised $h$-transform.
\newblock In \emph{Advances in Neural Information Processing Systems},
  volume~37, pp.\  19636--19682, 2024.

\bibitem[Dhariwal \& Nichol(2021)Dhariwal and Nichol]{dhariwal2021diffusion}
Prafulla Dhariwal and Alexander Nichol.
\newblock Diffusion models beat {GANs} on image synthesis.
\newblock In \emph{Advances in Neural Information Processing Systems},
  volume~34, pp.\  8780--8794, 2021.

\bibitem[Didi et~al.(2023)Didi, Vargas, Mathis, Dutordoir, Mathieu, Komorowska,
  and Lio]{didi2023framework}
Kieran Didi, Francisco Vargas, Simon Mathis, Vincent Dutordoir, Emile Mathieu,
  Urszula~Julia Komorowska, and Pietro Lio.
\newblock A framework for conditional diffusion modelling with applications in
  motif scaffolding for protein design.
\newblock arXiv preprint arXiv:2312.09236, 2023.

\bibitem[Doob(1957)]{doob1957}
J.~L. Doob.
\newblock Conditional {B}rownian motion and the boundary limits of harmonic
  functions.
\newblock \emph{Bulletin de la Soci{\'e}t{\'e} Math{\'e}matique de France},
  85:\penalty0 431--458, 1957.

\bibitem[Efron(2011)]{efron2011tweedie}
Bradley Efron.
\newblock {Tweedie's} formula and selection bias.
\newblock \emph{Journal of the American Statistical Association}, 106\penalty0
  (496):\penalty0 1602--1614, 2011.

\bibitem[Eyring et~al.(2025)Eyring, Karthik, Dosovitskiy, Ruiz, and
  Akata]{eyring2025hypernoise}
Luca Eyring, Shyamgopal Karthik, Alexey Dosovitskiy, Nataniel Ruiz, and Zeynep
  Akata.
\newblock Noise hypernetworks: Amortizing test-time compute in diffusion
  models.
\newblock In \emph{Advances in Neural Information Processing Systems}, 2025.

\bibitem[Go et~al.(2026)Go, Chung, Truong, Bhat, Mi, An, Zhao, Narnhofer,
  Belongie, Tombari, and Schindler]{go2026stitchvm}
Hyojun Go, Hyungjin Chung, Prune Truong, Goutam Bhat, Li~Mi, Zhaochong An,
  Zixiang Zhao, Dominik Narnhofer, Serge Belongie, Federico Tombari, and Konrad
  Schindler.
\newblock Stitched value model for diffusion alignment.
\newblock In \emph{Advances in Neural Information Processing Systems}, 2026.

\bibitem[Guo et~al.(2026)Guo, Tang, and Xu]{guo2026hardconstraint}
Zhengyi Guo, Wenpin Tang, and Renyuan Xu.
\newblock Conditional diffusion guidance under hard constraint: A stochastic
  analysis approach.
\newblock \emph{arXiv preprint arXiv:2602.05533}, 2026.

\bibitem[Haussmann \& Pardoux(1986)Haussmann and Pardoux]{haussmann1986time}
U.~G. Haussmann and E.~Pardoux.
\newblock Time reversal of diffusions.
\newblock \emph{The Annals of Probability}, 14\penalty0 (4):\penalty0
  1188--1205, 1986.

\bibitem[Heusel et~al.(2017)Heusel, Ramsauer, Unterthiner, Nessler, and
  Hochreiter]{heusel2017fid}
Martin Heusel, Hubert Ramsauer, Thomas Unterthiner, Bernhard Nessler, and Sepp
  Hochreiter.
\newblock {GANs} trained by a two time-scale update rule converge to a local
  {N}ash equilibrium.
\newblock In \emph{Advances in Neural Information Processing Systems},
  volume~30, 2017.

\bibitem[Ho \& Salimans(2022)Ho and Salimans]{ho2022classifierfree}
Jonathan Ho and Tim Salimans.
\newblock Classifier-free diffusion guidance.
\newblock arXiv preprint arXiv:2207.12598, 2022.
\newblock A short version appeared at the NeurIPS 2021 Workshop on Deep
  Generative Models and Downstream Applications.

\bibitem[Ho et~al.(2020)Ho, Jain, and Abbeel]{ho2020ddpm}
Jonathan Ho, Ajay~N. Jain, and Pieter Abbeel.
\newblock Denoising diffusion probabilistic models.
\newblock In \emph{Advances in Neural Information Processing Systems},
  volume~33, pp.\  6840--6851, 2020.

\bibitem[Hong et~al.(2023)Hong, Lee, Jang, and Kim]{hong2023sag}
Susung Hong, Gyuseong Lee, Wooseok Jang, and Seungryong Kim.
\newblock Improving sample quality of diffusion models using self-attention
  guidance.
\newblock In \emph{IEEE/CVF International Conference on Computer Vision}, pp.\
  7462--7471, 2023.

\bibitem[Janner et~al.(2022)Janner, Du, Tenenbaum, and
  Levine]{janner2022diffuser}
Michael Janner, Yilun Du, Joshua~B. Tenenbaum, and Sergey Levine.
\newblock Planning with diffusion for flexible behavior synthesis.
\newblock In \emph{Proceedings of the 39th International Conference on Machine
  Learning}, volume 162 of \emph{Proceedings of Machine Learning Research},
  pp.\  9902--9915, 2022.

\bibitem[Jensen et~al.(2026)Jensen, Schaufelberger, De~Santi, Jorner, and
  Krause]{jensen2026value}
Cristian~Perez Jensen, Luca Schaufelberger, Riccardo De~Santi, Kjell Jorner,
  and Andreas Krause.
\newblock Value matching: Scalable and gradient-free reward-guided flow
  adaptation.
\newblock In \emph{International Conference on Learning Representations}, 2026.

\bibitem[Jo et~al.(2022)Jo, Lee, and Hwang]{jo2022gdss}
Jaehyeong Jo, Seul Lee, and Sung~Ju Hwang.
\newblock Score-based generative modeling of graphs via the system of
  stochastic differential equations.
\newblock In \emph{Proceedings of the 39th International Conference on Machine
  Learning}, volume 162 of \emph{Proceedings of Machine Learning Research},
  pp.\  10362--10383, 2022.

\bibitem[Kim et~al.(2026)Kim, Yoon, Phunyaphibarn, Kim, Mardani, and
  Sung]{kim2026cdm}
Jaihoon Kim, Taehoon Yoon, Prin Phunyaphibarn, Seungjun Kim, Morteza Mardani,
  and Minhyuk Sung.
\newblock Contrastive distribution matching for amortized sequential {Monte
  Carlo} in discrete diffusion.
\newblock In \emph{Advances in Neural Information Processing Systems}, 2026.

\bibitem[Krizhevsky(2009)]{krizhevsky2009cifar}
Alex Krizhevsky.
\newblock Learning multiple layers of features from tiny images.
\newblock Technical report, University of Toronto, 2009.

\bibitem[Kynk{\"a}{\"a}nniemi et~al.(2024)Kynk{\"a}{\"a}nniemi, Aittala,
  Karras, Laine, Aila, and Lehtinen]{kynkaanniemi2024interval}
Tuomas Kynk{\"a}{\"a}nniemi, Miika Aittala, Tero Karras, Samuli Laine, Timo
  Aila, and Jaakko Lehtinen.
\newblock Applying guidance in a limited interval improves sample and
  distribution quality in diffusion models.
\newblock In \emph{Advances in Neural Information Processing Systems},
  volume~37, pp.\  122458--122483, 2024.

\bibitem[Lawson et~al.(2022)Lawson, Ravent\'{o}s, Warrington, and
  Linderman]{lawson2022sixo}
Dieterich Lawson, Allan Ravent\'{o}s, Andrew Warrington, and Scott Linderman.
\newblock {SIXO}: Smoothing inference with twisted objectives.
\newblock In \emph{Advances in Neural Information Processing Systems},
  volume~35, pp.\  38844--38858, 2022.

\bibitem[LeCun et~al.(1998)LeCun, Bottou, Bengio, and Haffner]{lecun1998mnist}
Yann LeCun, L{\'e}on Bottou, Yoshua Bengio, and Patrick Haffner.
\newblock Gradient-based learning applied to document recognition.
\newblock \emph{Proceedings of the IEEE}, 86\penalty0 (11):\penalty0
  2278--2324, 1998.

\bibitem[Lee et~al.(2023)Lee, Jo, and Hwang]{lee2023mood}
Seul Lee, Jaehyeong Jo, and Sung~Ju Hwang.
\newblock Exploring chemical space with score-based out-of-distribution
  generation.
\newblock In \emph{Proceedings of the 40th International Conference on Machine
  Learning}, volume 202 of \emph{Proceedings of Machine Learning Research},
  pp.\  18872--18892, 2023.

\bibitem[Li et~al.(2025)Li, Zhao, Wang, Scalia, Eraslan, Nair, Biancalani, Ji,
  Regev, Levine, and Uehara]{li2024svdd}
Xiner Li, Yulai Zhao, Chenyu Wang, Gabriele Scalia, Gokcen Eraslan, Surag Nair,
  Tommaso Biancalani, Shuiwang Ji, Aviv Regev, Sergey Levine, and Masatoshi
  Uehara.
\newblock Derivative-free guidance in continuous and discrete diffusion models
  with soft value-based decoding.
\newblock In \emph{Advances in Neural Information Processing Systems},
  volume~38, pp.\  95507--95545, 2025.

\bibitem[Lin et~al.(2014)Lin, Maire, Belongie, Hays, Perona, Ramanan,
  Doll{\'a}r, and Zitnick]{lin2014coco}
Tsung-Yi Lin, Michael Maire, Serge Belongie, James Hays, Pietro Perona, Deva
  Ramanan, Piotr Doll{\'a}r, and C.~Lawrence Zitnick.
\newblock Microsoft {COCO}: Common objects in context.
\newblock In \emph{Computer Vision -- ECCV 2014}, volume 8693 of \emph{Lecture
  Notes in Computer Science}, pp.\  740--755. Springer, 2014.

\bibitem[Lipinski et~al.(1997)Lipinski, Lombardo, Dominy, and
  Feeney]{lipinski1997}
Christopher~A. Lipinski, Franco Lombardo, Beryl~W. Dominy, and Paul~J. Feeney.
\newblock Experimental and computational approaches to estimate solubility and
  permeability in drug discovery and development settings.
\newblock \emph{Advanced Drug Delivery Reviews}, 23\penalty0 (1--3):\penalty0
  3--25, 1997.

\bibitem[Liu et~al.(2025)Liu, Liu, Liang, Li, Liu, Wang, Wan, Zhang, and
  Ouyang]{liu2025flowgrpo}
Jie Liu, Gongye Liu, Jiajun Liang, Yangguang Li, Jiaheng Liu, Xintao Wang,
  Pengfei Wan, Di~Zhang, and Wanli Ouyang.
\newblock Flow-{GRPO}: Training flow matching models via online {RL}.
\newblock In \emph{Advances in Neural Information Processing Systems}, 2025.

\bibitem[Ma et~al.(2024)Ma, Hu, Wang, and Sun]{ma2024elucidating}
Jiajun Ma, Tianyang Hu, Wenjia Wang, and Jiacheng Sun.
\newblock Elucidating the design space of classifier-guided diffusion
  generation.
\newblock In \emph{International Conference on Learning Representations}, 2024.

\bibitem[Moreira et~al.(2026)Moreira, Marques, Costeira, and
  Xiong]{moreira2026spectral}
Gabriel Moreira, Manuel Marques, Jo{\~a}o~Paulo Costeira, and Chenyan Xiong.
\newblock Spectral guidance for flexible and efficient control of diffusion
  models.
\newblock In \emph{Proceedings of the 43rd International Conference on Machine
  Learning}, volume 306 of \emph{Proceedings of Machine Learning Research},
  pp.\  90295--90322, 2026.

\bibitem[Mudgal et~al.(2024)Mudgal, Lee, Ganapathy, Li, Wang, Huang, Chen,
  Cheng, Collins, Strohman, Chen, Beutel, and Beirami]{mudgal2024controlled}
Sidharth Mudgal, Jong Lee, Harish Ganapathy, Yaguang Li, Tao Wang, Yanping
  Huang, Zhifeng Chen, Heng-Tze Cheng, Michael Collins, Trevor Strohman, Jilin
  Chen, Alex Beutel, and Ahmad Beirami.
\newblock Controlled decoding from language models.
\newblock In \emph{Proceedings of the 41st International Conference on Machine
  Learning}, volume 235 of \emph{Proceedings of Machine Learning Research},
  pp.\  36486--36503, 2024.

\bibitem[Nguyen et~al.(2025)Nguyen, Do, Kieu, and Nguyen]{nguyen2025hedit}
Toan Nguyen, Kien Do, Duc Kieu, and Thin Nguyen.
\newblock {h-Edit}: Effective and flexible diffusion-based editing via {Doob's}
  h-transform.
\newblock In \emph{Proceedings of the IEEE/CVF Conference on Computer Vision
  and Pattern Recognition (CVPR)}, pp.\  28490--28501, 2025.

\bibitem[Obukhov et~al.(2020)Obukhov, Seitzer, Wu, Zhydenko, Kyl, and
  Lin]{obukhov2020torchfidelity}
Anton Obukhov, Maximilian Seitzer, Po-Wei Wu, Semen Zhydenko, Jonathan Kyl, and
  Elvis Yu-Jing Lin.
\newblock High-fidelity performance metrics for generative models in {PyTorch}.
\newblock \url{https://github.com/toshas/torch-fidelity}, 2020.

\bibitem[{\O}ksendal(2003)]{oksendal2003sde}
Bernt {\O}ksendal.
\newblock \emph{Stochastic Differential Equations: An Introduction with
  Applications}.
\newblock Universitext. Springer, sixth edition, 2003.

\bibitem[Radford et~al.(2021)Radford, Kim, Hallacy, Ramesh, Goh, Agarwal,
  Sastry, Askell, Mishkin, Clark, Krueger, and Sutskever]{radford2021clip}
Alec Radford, Jong~Wook Kim, Chris Hallacy, Aditya Ramesh, Gabriel Goh,
  Sandhini Agarwal, Girish Sastry, Amanda Askell, Pamela Mishkin, Jack Clark,
  Gretchen Krueger, and Ilya Sutskever.
\newblock Learning transferable visual models from natural language
  supervision.
\newblock In \emph{Proceedings of the 38th International Conference on Machine
  Learning}, volume 139 of \emph{Proceedings of Machine Learning Research},
  pp.\  8748--8763, 2021.

\bibitem[Ramakrishnan et~al.(2014)Ramakrishnan, Dral, Rupp, and von
  Lilienfeld]{ramakrishnan2014qm9}
Raghunathan Ramakrishnan, Pavlo~O. Dral, Matthias Rupp, and O.~Anatole von
  Lilienfeld.
\newblock Quantum chemistry structures and properties of 134 kilo molecules.
\newblock \emph{Scientific Data}, 1:\penalty0 140022, 2014.

\bibitem[Rombach et~al.(2022)Rombach, Blattmann, Lorenz, Esser, and
  Ommer]{rombach2022ldm}
Robin Rombach, Andreas Blattmann, Dominik Lorenz, Patrick Esser, and Bj{\"o}rn
  Ommer.
\newblock High-resolution image synthesis with latent diffusion models.
\newblock In \emph{Proceedings of the IEEE/CVF Conference on Computer Vision
  and Pattern Recognition (CVPR)}, pp.\  10684--10695, 2022.

\bibitem[Sahu et~al.(2026)Sahu, Banerjee, and Wu]{sahu2026reliable}
Sharan Sahu, Arisina Banerjee, and Yuchen Wu.
\newblock Provably reliable classifier guidance via cross-entropy control.
\newblock In \emph{Advances in Neural Information Processing Systems}, 2026.

\bibitem[Song et~al.(2021{\natexlab{a}})Song, Meng, and Ermon]{song2021ddim}
Jiaming Song, Chenlin Meng, and Stefano Ermon.
\newblock Denoising diffusion implicit models.
\newblock In \emph{International Conference on Learning Representations},
  2021{\natexlab{a}}.

\bibitem[Song et~al.(2023)Song, Vahdat, Mardani, and Kautz]{song2023pigdm}
Jiaming Song, Arash Vahdat, Morteza Mardani, and Jan Kautz.
\newblock Pseudoinverse-guided diffusion models for inverse problems.
\newblock In \emph{International Conference on Learning Representations}, 2023.

\bibitem[Song et~al.(2021{\natexlab{b}})Song, Sohl-Dickstein, Kingma, Kumar,
  Ermon, and Poole]{song2021sde}
Yang Song, Jascha Sohl-Dickstein, Diederik~P. Kingma, Abhishek Kumar, Stefano
  Ermon, and Ben Poole.
\newblock Score-based generative modeling through stochastic differential
  equations.
\newblock In \emph{International Conference on Learning Representations},
  2021{\natexlab{b}}.

\bibitem[Uehara et~al.(2024)Uehara, Zhao, Biancalani, and
  Levine]{uehara2024understanding}
Masatoshi Uehara, Yulai Zhao, Tommaso Biancalani, and Sergey Levine.
\newblock Understanding reinforcement learning-based fine-tuning of diffusion
  models: A tutorial and review.
\newblock arXiv preprint arXiv:2407.13734, 2024.

\bibitem[Vignac et~al.(2023)Vignac, Krawczuk, Siraudin, Wang, Cevher, and
  Frossard]{vignac2023digress}
Cl{\'e}ment Vignac, Igor Krawczuk, Antoine Siraudin, Bohan Wang, Volkan Cevher,
  and Pascal Frossard.
\newblock {DiGress}: Discrete denoising diffusion for graph generation.
\newblock In \emph{International Conference on Learning Representations}, 2023.

\bibitem[Wilson(1927)]{wilson1927}
Edwin~B. Wilson.
\newblock Probable inference, the law of succession, and statistical inference.
\newblock \emph{Journal of the American Statistical Association}, 22\penalty0
  (158):\penalty0 209--212, 1927.

\bibitem[Xu et~al.(2023)Xu, Liu, Wu, Tong, Li, Ding, Tang, and
  Dong]{xu2023imagereward}
Jiazheng Xu, Xiao Liu, Yuchen Wu, Yuxuan Tong, Qinkai Li, Ming Ding, Jie Tang,
  and Yuxiao Dong.
\newblock {ImageReward}: Learning and evaluating human preferences for
  text-to-image generation.
\newblock In \emph{Advances in Neural Information Processing Systems}, 2023.

\bibitem[Yang \& Klein(2021)Yang and Klein]{yang2021fudge}
Kevin Yang and Dan Klein.
\newblock {FUDGE}: Controlled text generation with future discriminators.
\newblock In \emph{Proceedings of the 2021 Conference of the North American
  Chapter of the Association for Computational Linguistics: Human Language
  Technologies}, pp.\  3511--3535, 2021.

\bibitem[Yang et~al.(2024)Yang, Ding, Cai, Yu, Wang, and Shi]{yang2024dsg}
Lingxiao Yang, Shutong Ding, Yifan Cai, Jingyi Yu, Jingya Wang, and Ye~Shi.
\newblock Guidance with spherical {G}aussian constraint for conditional
  diffusion.
\newblock In \emph{Proceedings of the 41st International Conference on Machine
  Learning}, volume 235 of \emph{Proceedings of Machine Learning Research},
  pp.\  56071--56095, 2024.

\bibitem[Yin et~al.(2024)Yin, Gharbi, Zhang, Shechtman, Durand, Freeman, and
  Park]{yin2024dmd}
Tianwei Yin, Micha{\"e}l Gharbi, Richard Zhang, Eli Shechtman, Fr{\'e}do
  Durand, William~T. Freeman, and Taesung Park.
\newblock One-step diffusion with distribution matching distillation.
\newblock In \emph{IEEE/CVF Conference on Computer Vision and Pattern
  Recognition}, 2024.

\bibitem[Yu et~al.(2023)Yu, Wang, Zhao, Ghanem, and Zhang]{yu2023freedom}
Jiwen Yu, Yinhuai Wang, Chen Zhao, Bernard Ghanem, and Jian Zhang.
\newblock {FreeDoM}: Training-free energy-guided conditional diffusion model.
\newblock In \emph{Proceedings of the IEEE/CVF International Conference on
  Computer Vision (ICCV)}, pp.\  23174--23184, 2023.

\bibitem[Zhao et~al.(2024)Zhao, Brekelmans, Makhzani, and
  Grosse]{zhao2024probabilistic}
Stephen Zhao, Rob Brekelmans, Alireza Makhzani, and Roger~Baker Grosse.
\newblock Probabilistic inference in language models via twisted sequential
  {Monte Carlo}.
\newblock In \emph{Proceedings of the 41st International Conference on Machine
  Learning}, volume 235 of \emph{Proceedings of Machine Learning Research},
  pp.\  60704--60748, 2024.

\bibitem[Zheng et~al.(2025)Zheng, Chu, Wang, Kovachki, Baptista, and
  Yue]{zheng2025enkg}
Hongkai Zheng, Wenda Chu, Austin Wang, Nikola~B. Kovachki, Ricardo Baptista,
  and Yisong Yue.
\newblock Ensemble {Kalman} diffusion guidance: A derivative-free method for
  inverse problems.
\newblock \emph{Transactions on Machine Learning Research}, 2025.

\bibitem[Zheng et~al.(2026)Zheng, Chen, Ye, Wang, Zhang, Jiang, Su, Ermon, Zhu,
  and Liu]{zheng2026diffusionnft}
Kaiwen Zheng, Huayu Chen, Haotian Ye, Haoxiang Wang, Qinsheng Zhang, Kai Jiang,
  Hang Su, Stefano Ermon, Jun Zhu, and Ming-Yu Liu.
\newblock {DiffusionNFT}: Online diffusion reinforcement with forward process.
\newblock In \emph{International Conference on Learning Representations}, 2026.

\bibitem[Zhu et~al.(2026)Zhu, Ye, Liu, Wang, and Chen]{zhu2026doit}
Qijie Zhu, Zeqi Ye, Han Liu, Zhaoran Wang, and Minshuo Chen.
\newblock Training-free adaptation of diffusion models via {Doob's}
  $h$-transform.
\newblock In \emph{International Conference on Machine Learning}, 2026.

\end{thebibliography}
\bibliographystyle{preprint}

\appendix
\raggedbottom

\section{Proofs}
\label{app:proofs}

\paragraph{Setting and standing assumption.}
Throughout, $(X_t)_{t\in[0,T]}$ solves $\mathrm{d}X_t=f(X_t,t)\,\mathrm{d}t+g(t)\,\mathrm{d}W_t$
with $X_0\sim p_0$ independent of $W$, the set $C$ is measurable with
$\bar h=\Pr(X_0\in C)\in(0,1)$, and $Y=\mathbf 1_C(X_0)$. We write $q_t(\cdot\mid x_0)$
for the transition density of $X_t$ given $X_0=x_0$, $p_t$ for the marginal density,
$\mathcal G_t=\sigma(X_r:r\ge t)$, and
\[
m_t(x)=\int_C q_t(x\mid x_0)\,p_0(\mathrm{d}x_0),\qquad h_t(x)=\frac{m_t(x)}{p_t(x)},
\]
so that $h_t(x)=\Pr(X_0\in C\mid X_t=x)$ by Bayes' rule, and $m_t/\bar h=p_t(\cdot\mid X_0\in C)$.
The regularity conditions of Theorem~\ref{thm:char} are the following.

\textbf{(A)} The forward SDE has a unique strong solution, $f(\cdot,t)$ is $C^1$ and $g>0$.
For $t\in(0,T]$, $p_t$ and $m_t$ are positive, of class $C^{1,2}$ in $(t,x)$,
and solve the forward equation $\partial_t\rho=\mathcal F_t\rho:=-\nabla\cdot(f\rho)+\tfrac12
g^2\Delta\rho$ (for $m_t$ this follows from the equation for $q_t(\cdot\mid x_0)$ by
integrating over $x_0\in C$). For $\rho=p$ and for $\rho=m/\bar h$, the reverse-time SDE
\begin{equation}
\mathrm{d}\bar X_s=\big[-f(\bar X_s,T-s)+g(T-s)^2\,\nabla\log\rho_{T-s}(\bar X_s)\big]\,\mathrm{d}s
+g(T-s)\,\mathrm{d}\bar W_s,\qquad s\in[0,T),
\label{eq:revsde}
\end{equation}
is well posed from every initial law, and its solution started from $\rho_T$ has the law of the
time reversal of the forward process started from $\rho_0$
\citep{anderson1982reverse,haussmann1986time}.

For the variance-preserving and variance-exploding processes used in this paper the forward
kernels are Gaussian, so $p_t$ and $m_t$ are smooth and strictly positive for $t>0$; (A) is the
usual assumption behind the reverse-time formulation of score-based models \citep{song2021sde}.

\subsection{Proof of Proposition~\ref{prop:target}}
\label{app:prop1}

\begin{proof}
Fix $t$ and let $h$ be measurable with $\mathbb E[h(X_t)^2]<\infty$. Expanding the square
around $h_t(X_t)$,
\begin{align*}
\mathbb E\big[(h(X_t)-Y)^2\big]&=\mathbb E\big[(h(X_t)-h_t(X_t))^2\big]
+\mathbb E\big[(h_t(X_t)-Y)^2\big]\\
&\quad+2\,\mathbb E\big[(h(X_t)-h_t(X_t))\,\mathbb E[h_t(X_t)-Y\mid X_t]\big].
\end{align*}
The last term vanishes because $\mathbb E[Y\mid X_t]=h_t(X_t)$, and the second does not
depend on $h$. The objective is therefore minimized exactly by the functions that agree with
$h_t$ $p_t$-almost everywhere, and by $h_t$ alone among continuous functions, since $p_t>0$.
$\mathcal L_{\mathrm{sup}}$ averages this objective over noise levels drawn uniformly from the
time grid, so its minimizer is $h_t$ at every level. Only pairs $(X_0,X_t)$ enter, and the
forward kernel generates them in closed form.
\end{proof}

\subsection{Proof of Theorem~\ref{thm:char}}
\label{app:thm1}

The initial condition $h_0=\mathbf 1_C$ is discontinuous, so we state in which sense it is
attained: along the process. Theorem~\ref{thm:char} is the combination of the four parts below.

\begin{theoremprecise}
Assume (A). Then
(a)~$h\in C^{1,2}((0,T]\times\mathbb R^d)$, $0\le h\le 1$, and $\partial_t h=\widehat{\mathcal L}_th$
on $(0,T]\times\mathbb R^d$;
(b)~$h_t(X_t)\to\mathbf 1_C(X_0)$ almost surely and in $L^2$ as $t\downarrow0$;
(c)~if $u\in C^{1,2}((0,T]\times\mathbb R^d)$ is bounded, satisfies $\partial_tu=\widehat{\mathcal
L}_tu$, and $u(\varepsilon,X_\varepsilon)\to\mathbf 1_C(X_0)$ in $L^1$ as $\varepsilon\downarrow0$,
then $u=h$;
(d)~the reverse-time dynamics with drift augmented by $g(t)^2\nabla\log h_t$, started from
$p_T(\cdot\mid X_0\in C)$, have marginal $p_t(\cdot\mid X_0\in C)$ at every $t\in(0,T]$, and
their limit as $t\downarrow0$ exists almost surely and has law $p_0(\cdot\mid X_0\in C)$.
\end{theoremprecise}

\begin{proof}
\emph{(a) The equation.} By (A), $h=m/p$ is of class $C^{1,2}$ with values in $[0,1]$. For
$C^{1,2}$ functions the product rule gives
\[
\mathcal F_t(hp)=h\,\mathcal F_tp-p\,f\cdot\nabla h+g^2\,\nabla p\cdot\nabla h+\tfrac12g^2p\,\Delta h .
\]
Since $\partial_tm=\mathcal F_tm$ and $\partial_tp=\mathcal F_tp$,
\[
\partial_th=\frac{\partial_tm-h\,\partial_tp}{p}=\frac{\mathcal F_t(hp)-h\,\mathcal F_tp}{p}
=\tfrac12g^2\Delta h+\big(g^2\nabla\log p_t-f\big)\cdot\nabla h=\widehat{\mathcal L}_th .
\]
The same computation with $\int\phi(x_0)\,q_t(\cdot\mid x_0)\,p_0(\mathrm dx_0)$ in place of
$m_t$ shows that $u(x,t)=\mathbb E[\phi(X_0)\mid X_t=x]$ solves $\partial_tu=\widehat{\mathcal
L}_tu$ for bounded $\phi$ with the corresponding regularity, as stated in
Section~\ref{sec:background}.

\emph{(b) The initial condition.} Because $(X_t)$ is Markov, $X_0$ is conditionally
independent of $(X_r)_{r>t}$ given $X_t$, so $h_t(X_t)=\mathbb E[Y\mid X_t]=\mathbb E[Y\mid
\mathcal G_t]$. As $t\downarrow0$ the $\sigma$-fields $\mathcal G_t$ increase to
$\mathcal G_{0+}=\bigvee_{t>0}\mathcal G_t$, which contains $\sigma(X_0)$ because
$X_0=\lim_{t\downarrow0}X_t$ by continuity of paths. By (a) and continuity of paths,
$s\mapsto h_{T-s}(X_{T-s})=\mathbb E[Y\mid\mathcal G_{T-s}]$ is a continuous bounded martingale
for the increasing filtration $(\mathcal G_{T-s})_{s\in[0,T)}$, so it converges almost surely as
$s\uparrow T$. Along any sequence $t_n\downarrow0$, L\'evy's upward theorem identifies the
limit as $\mathbb E[Y\mid\mathcal G_{0+}]=Y$. Hence $h_t(X_t)\to Y$ almost surely, and in $L^2$
by bounded convergence.

\emph{(c) Uniqueness.} Let $\bar X_s=X_{T-s}$ and $v(s,x)=u(T-s,x)$ for $s\in[0,T)$. By (A),
$\bar X$ solves \eqref{eq:revsde} with $\rho=p$, whose generator at time $s$ is
$\widehat{\mathcal L}_{T-s}$, and $\partial_sv=-\widehat{\mathcal L}_{T-s}v$. By It\^o's formula
\citep{oksendal2003sde}, $v(s,\bar X_s)$ is a local martingale on $[0,T-\varepsilon]$. It is
bounded, hence a martingale, also with respect to the filtration
$\sigma(\bar X_r:r\le s)=\mathcal G_{T-s}$ generated by $\bar X$; this property depends only on
the law of $\bar X$, so it holds for the time reversal of $X$. Therefore
$u(t,X_t)=\mathbb E[u(\varepsilon,X_\varepsilon)\mid\mathcal G_t]$ for $0<\varepsilon<t$.
Letting $\varepsilon\downarrow0$ and using that conditional expectation is a contraction in
$L^1$, $u(t,X_t)=\mathbb E[Y\mid\mathcal G_t]=h_t(X_t)$ almost surely. Hence $u(t,\cdot)=h_t$
$p_t$-almost everywhere, and everywhere because both are continuous and $p_t>0$.

\emph{(d) Marginals.} Conditionally on $X_0\in C$, the process solves the same forward SDE
(the Brownian motion is independent of $X_0$) from the initial law
$p_0(\cdot\mid X_0\in C)=\mathbf 1_Cp_0/\bar h$, so its marginals are
$m_t/\bar h=p_t\,h_t/\bar h=p_t(\cdot\mid X_0\in C)$. By (A), its time reversal solves
\eqref{eq:revsde} with $\rho=m/\bar h$, whose drift contains
$g^2\nabla\log(m_t/\bar h)=g^2\nabla\log p_t+g^2\nabla\log h_t$: these are the reverse-time
dynamics with drift augmented by $g(t)^2\nabla\log h_t$. Started from
$p_T(\cdot\mid X_0\in C)$, they have marginal $p_t(\cdot\mid X_0\in C)$ at every $t\in(0,T]$.
The time reversal of the conditioned forward process extends continuously to $s=T$ with value
$X_0$; since the solution has the same law on $C([0,T),\mathbb R^d)$, its limit as $s\uparrow T$
exists almost surely and has law $p_0(\cdot\mid X_0\in C)$.
\end{proof}

\paragraph{Initialization.}
Samplers start the reverse dynamics from $p_T$, or its Gaussian approximation, rather than from
$p_T(\cdot\mid X_0\in C)$. Since $p_T(\cdot\mid X_0\in C)/p_T=h_T/\bar h$, the bound
$\mathbb E|h_T(X_T)-\bar h|\le\sqrt{I(X_0;X_T)/2}$ from the proof of
Proposition~\ref{prop:contraction}(ii) gives
\[
\mathrm{TV}\big(p_T(\cdot\mid X_0\in C),\,p_T\big)=\frac{\mathbb E\,|h_T(X_T)-\bar h|}{2\bar h}
\le\frac{\sqrt{I(X_0;X_T)/2}}{2\bar h},
\]
and the guided reverse dynamics, being a Markov kernel, cannot increase this distance; starting
from $\mathcal N(0,I)$ instead of $p_T$ adds $\mathrm{TV}(p_T,\mathcal N(0,I))$ by the triangle
inequality. The error due to initialization therefore vanishes as the forward process forgets
$X_0$.

\subsection{Proof of Proposition~\ref{prop:contraction}}
\label{app:prop2}

\begin{proof}
(i) $\mathbb E[h_t(X_t)]=\mathbb E\big[\mathbb E[Y\mid X_t]\big]=\mathbb E[Y]=\bar h$. For
$0\le s<t$, $X_0\to X_s\to X_t$ is a Markov chain, so
$\mathbb E[Y\mid X_s,X_t]=\mathbb E[Y\mid X_s]=h_s(X_s)$ and, by the tower property,
$h_t(X_t)=\mathbb E[h_s(X_s)\mid X_t]$. The law of total variance gives
$\sigma^2(s)=\sigma^2(t)+\mathbb E\big[\mathrm{Var}(h_s(X_s)\mid X_t)\big]\ge\sigma^2(t)$.

(ii) Since $h_t(X_t)$ and $\bar h$ lie in $[0,1]$, $\sigma^2(t)\le\mathbb E|h_t(X_t)-\bar h|$.
The joint law of $(Y,X_t)$ has density $h_tp_t$ at $Y=1$ and $(1-h_t)p_t$ at $Y=0$, and the
product of its marginals has densities $\bar h\,p_t$ and $(1-\bar h)\,p_t$. Hence
\[
\mathrm{TV}\big(P_{Y,X_t},\,P_Y\otimes P_{X_t}\big)
=\tfrac12\int\big(|h_t-\bar h|+|(1-h_t)-(1-\bar h)|\big)\,p_t
=\mathbb E\,|h_t(X_t)-\bar h| .
\]
Pinsker's inequality bounds this by
$\sqrt{\mathrm{KL}(P_{Y,X_t}\,\|\,P_Y\otimes P_{X_t})/2}=\sqrt{I(Y;X_t)/2}$, and the
data-processing inequality for $Y\to X_0\to X_t$ gives $I(Y;X_t)\le I(X_0;X_t)$
\citep{cover2006elements}.

(iii) For the variance-preserving process,
$X_t=\sqrt{\bar\alpha_t}\,X_0+\sqrt{1-\bar\alpha_t}\,\varepsilon$ with
$\varepsilon\sim\mathcal N(0,I_d)$ independent of $X_0$. Writing $H$ for differential
entropy, $I(X_0;X_t)=H(X_t)-H(X_t\mid X_0)$ with
$H(X_t\mid X_0)=\tfrac d2\log\big(2\pi e(1-\bar\alpha_t)\big)$. Among densities on
$\mathbb R^d$ with second moment $s_t=\mathbb E\|X_t\|^2=\bar\alpha_tm_2+(1-\bar\alpha_t)d$,
the isotropic Gaussian has the largest entropy \citep{cover2006elements}, so
$H(X_t)\le\tfrac d2\log(2\pi e\,s_t/d)$. Subtracting,
\[
I(X_0;X_t)\le\frac d2\log\frac{\bar\alpha_tm_2+(1-\bar\alpha_t)d}{(1-\bar\alpha_t)d}
=\frac d2\log\Big(1+\frac{\bar\alpha_t\,m_2}{(1-\bar\alpha_t)\,d}\Big),
\]
which tends to $0$ as $\bar\alpha_t\to0$; by (ii), so does $\sigma^2(t)$.
\end{proof}

\subsection{Proof of Proposition~\ref{prop:estimand}}
\label{app:prop3}

Let $\hat X$ be the model's reverse-time process, \eqref{eq:revsde} with $\nabla\log\rho_{T-s}$
replaced by $s_\theta(\cdot,T-s)$, and write $\hat X_{(t)}$ for its state at diffusion time $t$.
A rollout-based estimator regresses $\mathbf 1_C(\hat X_{(0)})$ on the state $\hat X_{(t)}=x$
from which the rollout starts, so it targets
$h^{\mathrm{model}}_t(x)=\Pr(\hat X_{(0)}\in C\mid\hat X_{(t)}=x)$. We assume that this SDE
is well posed on $[0,T]$ and that $h^{\mathrm{model}}\in C^{1,2}((0,T]\times\mathbb R^d)$; for the
size of the gap we also assume the integrability condition stated below and that both
$h^{\mathrm{data}}$ and $h^{\mathrm{model}}$ approach $\mathbf 1_C$ along the model's paths.

\begin{proof}
\emph{Equation.} By the Markov property,
$h^{\mathrm{model}}_t(x)=\mathbb E\big[h^{\mathrm{model}}_\varepsilon(\hat X_{(\varepsilon)})
\mid\hat X_{(t)}=x\big]$ for $0<\varepsilon<t$, so $h^{\mathrm{model}}$ evaluated along $\hat X$
is a martingale and, by It\^o's formula, solves the Kolmogorov backward equation of $\hat X$
\citep{oksendal2003sde}. In diffusion time this equation reads
$\partial_th^{\mathrm{model}}=\widehat{\mathcal L}^\theta_th^{\mathrm{model}}$ with
$\widehat{\mathcal L}^\theta_t=\tfrac12g^2\Delta+(g^2s_\theta-f)\cdot\nabla$, which
is~\eqref{eq:cauchy} with $\nabla\log p_t$ replaced by $s_\theta$. The initial condition is
attained along $\hat X$, by the argument of Theorem~\ref{thm:char}(b).

\emph{Equality.} If $s_\theta=\nabla\log p_t$, the two Cauchy problems and the two reverse
processes coincide: started from $p_T$, the model process has, by (A), the law of the time
reversal of $X$, so attainment of the initial condition along $\hat X$ is attainment along $X$,
and part~(c) of Theorem~\ref{thm:char} gives $h^{\mathrm{model}}=h^{\mathrm{data}}$ for every
$C$. More generally, if the model's reverse
process has the transition laws of the true reverse process, then
$\Pr(\hat X_{(0)}\in C\mid\hat X_{(t)}=x)=\Pr(X_0\in C\mid X_t=x)$ by definition.

\emph{Size of the gap.} Let $e_t=\nabla\log p_t-s_\theta(\cdot,t)$ and
$w_t=h^{\mathrm{data}}_t-h^{\mathrm{model}}_t$. Subtracting the two equations,
$\partial_tw=\widehat{\mathcal L}^\theta_tw+g(t)^2\,e_t\cdot\nabla h^{\mathrm{data}}_t$.
Applying It\^o's formula to $w$ along $\hat X$ from diffusion time $t$ down to $\varepsilon$,
and assuming
$\mathbb E\big[\int_\varepsilon^tg(\tau)^2\,|e_\tau\cdot\nabla h^{\mathrm{data}}_\tau|(\hat
X_{(\tau)})\,\mathrm d\tau\mid\hat X_{(t)}=x\big]<\infty$ (with $|w|\le1$ this also makes the
stochastic integral a true martingale),
\[
w_t(x)=\mathbb E\big[w_\varepsilon(\hat X_{(\varepsilon)})\,\big|\,\hat X_{(t)}=x\big]
+\mathbb E\Big[\int_\varepsilon^t g(\tau)^2\,e_\tau(\hat X_{(\tau)})\cdot
\nabla h^{\mathrm{data}}_\tau(\hat X_{(\tau)})\,\mathrm{d}\tau\;\Big|\;\hat X_{(t)}=x\Big].
\]
The first term vanishes as $\varepsilon\downarrow0$ when both functions approach
$\mathbf 1_C$ along the model's paths, and the gap is then an integral of the score error along the
model's paths, weighted by the gradient of $h^{\mathrm{data}}$: it vanishes when $e\equiv0$ and
does not depend on how either function is estimated.
\end{proof}

\subsection{The closed form of the guidance field}
\label{app:tweedie}

For the variance-preserving kernel
$q_t(x\mid x_0)=\mathcal N\big(x;\sqrt{\bar\alpha_t}\,x_0,(1-\bar\alpha_t)I\big)$,
$\nabla_xq_t(x\mid x_0)=q_t(x\mid x_0)\,(\sqrt{\bar\alpha_t}\,x_0-x)/(1-\bar\alpha_t)$.
Differentiating $m_t$ and $p_t$ under the integral sign, which the Gaussian factor
permits,
\[
\nabla\log m_t(x)=\frac{\sqrt{\bar\alpha_t}\,\mathbb E[X_0\mid X_t=x,\,X_0\in C]-x}{1-\bar\alpha_t},
\quad
\nabla\log p_t(x)=\frac{\sqrt{\bar\alpha_t}\,\mathbb E[X_0\mid X_t=x]-x}{1-\bar\alpha_t},
\]
the second being Tweedie's formula \citep{efron2011tweedie}. Their difference is
\eqref{eq:tweedie}. The reference field of Tables~\ref{tab:field} and~\ref{tab:e1} estimates
both conditional means from the same reverse trajectories of the frozen model started at
$x_t$. These trajectories follow the model's reverse process, so the reference is the right-hand
side of \eqref{eq:tweedie} with the model's posterior in place of the data posterior. It equals
$\nabla\log h^{\mathrm{model}}_t$ when the model's reverse kernel from level $t$ to $0$ has the
form $K_t(\mathrm dx_0\mid x)\propto\mu_t(\mathrm dx_0)\,q_t(x\mid x_0)$, as for an exact score,
and otherwise differs from it by a term due to the model's error. Since $h_\theta$ estimates
$h^{\mathrm{data}}_t$, the reported cosines contain the estimand gap of
Proposition~\ref{prop:estimand} and this discrepancy.

\subsection{The guided sampler and composition}
\label{app:sampler}

In the DDPM parameterization $s_\theta=-\epsilon_\theta/\sqrt{1-\bar\alpha_t}$, so the
replacement~\eqref{eq:sampler} is the same as replacing the score by
$\tilde s=s_\theta+\gamma\nabla\log(h_\theta+\delta)$. With $\gamma=1$, $\delta=0$ and exact
$s_\theta$ and $h_\theta$, $\tilde s=\nabla\log p_t+\nabla\log h_t=\nabla\log p_t(\cdot\mid
X_0\in C)$, the score in the drift of Theorem~\ref{thm:char}(d); a sampler that discretizes the
reverse-time dynamics with a score therefore discretizes the $h$-transformed dynamics when it is
given $\tilde s$. DDIM discretizes the probability-flow ODE, whose velocity uses the score in the
same way; with the conditional score it transports $p_T(\cdot\mid X_0\in C)$ through the same
marginals, which also solve the forward equation. On QM9 we add
$\gamma\nabla\log(h_\theta+\delta)$ to the node-feature score only, so even at $\gamma=1$ the
correction is partial. On Stable Diffusion the correction is added to the
classifier-free-guided noise prediction, so it conditions the guided sampler rather than a single
score model.

If, for every $t\in(0,T]$, $\mathbf 1_{C_1}(X_0)$ and $\mathbf 1_{C_2}(X_0)$ are
conditionally independent given $X_t$, then
$\Pr(X_0\in C_1\cap C_2\mid X_t=x)=h^{(1)}_t(x)\,h^{(2)}_t(x)$, the correction for the
intersection is $\nabla\log h^{(1)}_t+\nabla\log h^{(2)}_t$, and Theorem~\ref{thm:char}(d)
applied to $C_1\cap C_2$ gives marginals $p_t(\cdot\mid X_0\in C_1\cap C_2)$. With weights
$(w_1,w_2)$ the drift at time $t$ uses the score of the density proportional to
$p_t\,(h^{(1)}_t)^{w_1}(h^{(2)}_t)^{w_2}$; for weights other than $(1,1)$ this family does not in
general solve the forward equation, so it is not the marginal law of the guided process, and the
weights act as a trade-off between the constraints.

\section{Implementation details}
\label{app:details}

\subsection{Training objective}
\label{app:objective}

The heads of Sections~\ref{sec:cifar}, \ref{sec:mol} and~\ref{sec:sd} minimize
$\mathcal L_{\mathrm{bdry}}+\mathcal L_{\mathrm{sup}}+0.1\,\mathcal L_{\mathrm{fc}}$ exactly as
written in Section~\ref{sec:method}: $t$ is drawn uniformly from $\{0,\dots,T-1\}$ for
$\mathcal L_{\mathrm{sup}}$ and from $\{0,\dots,T-2\}$ for $\mathcal L_{\mathrm{fc}}$, and
$x_t$ and $x'_{t+1}$ are noised from the same $x_0$ with independent noise. (Noise levels are
indexed from $0$ to $T-1$ in code, matching the samplers' timestep indices; Algorithm~\ref{alg:main}
draws one $t$ for both terms, the code draws them independently, and the expected objective is
the same.) The two regularizers play different roles. $\mathcal L_{\mathrm{bdry}}$ is minimized
by $h(\cdot,0)=\mathbf 1_C$, the initial condition itself, so it adds weight at $t=0$ without
changing the target. $\mathcal L_{\mathrm{fc}}$ is a smoothing penalty and moves the target:
when its stop-gradient target equals $h_{t+1}$, the minimizer of
$\mathcal L_{\mathrm{sup}}+\lambda_{\mathrm{fc}}\mathcal L_{\mathrm{fc}}$ at level $t$ is, pointwise,
\[
\frac{h_t(x)+\lambda_{\mathrm{fc}}\,\mathbb E\big[h_{t+1}(X'_{t+1})\mid X_t=x\big]}{1+\lambda_{\mathrm{fc}}},
\]
a convex combination of $h_t$ and the posterior mean, given $X_t=x$, of the label after an extra
independent noising of $X_0$ to level $t+1$. The second term need not be close to $h_t$ even for
adjacent levels. With $\lambda_{\mathrm{fc}}=0.1$ the bias is at most $1/11$ in absolute value,
and Proposition~\ref{prop:target} rests on $\mathcal L_{\mathrm{sup}}$ alone.

The MNIST heads of Sections~\ref{sec:theory} and~\ref{sec:cost} and
Appendix~\ref{app:mnistcomp} (digit $3$, even digit, digit $>5$) use the same three terms and
weights with one difference in the consistency term: its pair is $(x_t,x_{t-1})$, noised from
the same $x_0$ with a shared noise draw, and the stop-gradient is on the less noisy member
$h_\theta(x_{t-1},t-1)$. The two members of this pair are close, so this term moves the target
less than independent noisings do; the small-$h$ evidence of Tables~\ref{tab:field}
and~\ref{tab:e1} comes from these heads.

\subsection{Evaluation protocol}
\label{app:protocol}

\emph{Satisfaction} is the fraction of generated samples that satisfy the constraint; every run
also records a Wilson 95\% interval \citep{wilson1927}. For the pixel predicates, the Stable
Diffusion predicate and the molecular predicates, the function that labels the training data
also scores the samples. For the MNIST digit constraints and the CIFAR-10 classes cat and ship,
the training labels are the dataset labels, and a classifier trained separately on the same
dataset judges the generated samples. Tables report means over seeds and, where shown, the
standard deviation over seeds (normalized by the number of seeds).

\emph{FID} \citep{heusel2017fid} is computed with \texttt{torch-fidelity}
\citep{obukhov2020torchfidelity} between the samples of one seed and a reference set; the
reference sets are listed per experiment below. The \emph{CLIP score} is $100$ times the cosine
similarity between the image and prompt embeddings of the OpenAI CLIP ViT-L/14 model
\citep{radford2021clip}, averaged over images.

\emph{Molecular metrics.} Validity is the fraction of generated graphs that decode to a valid
RDKit molecule. Satisfaction and uniqueness are computed over valid molecules, and novelty is the
fraction of unique valid molecules whose SMILES string is not among the 120{,}732 distinct SMILES
of the $120{,}803$ training molecules. QM9 is drawn from an exhaustive enumeration of small molecules,
so a novel molecule here is one absent from the training set, not new chemistry.

\emph{Timing} is wall-clock time per sample over a full run on one GPU. The MNIST, CIFAR-10 and
QM9 runs used an NVIDIA RTX~4090, except the QM9 SVDD-PM run and the unguided reference it is
compared with, which both ran on an NVIDIA T4, and the runs of Appendix~\ref{app:alloc},
Table~\ref{tab:cgstd} and the unguided Euler--Maruyama QM9 baseline, which used an NVIDIA A100. On Stable Diffusion, the unguided,
guided and classifier-guided runs of Tables~\ref{tab:sd} and~\ref{tab:sdfull} shared one
session, the SVDD-PM run used a second one, and the runs of Table~\ref{tab:sdstd} a third, on an
NVIDIA A100.

\subsection{MNIST}
\label{app:mnist}

\emph{Data and base model.} MNIST \citep{lecun1998mnist}, scaled to $[-1,1]$: the $60{,}000$
training images label and train the heads, and held-out quantities use the test set. The base
model is a DDPM with a small U-Net that we trained on MNIST, with a linear schedule from
$\beta_1=10^{-4}$ to $\beta_T=0.02$, $T=1000$, sampled with $1000$ ancestral steps.

\emph{Heads.} Every MNIST head, including the SVDD-MC value network, has $122{,}657$
parameters: three $3\times3$ convolutions with $32$, $64$ and $128$ channels (ReLU, two
$2\times2$ max-pools, global average pooling), a $32$-dimensional sinusoidal time embedding
followed by a linear layer, and a head $160\to128\to64\to1$ with a sigmoid output. The heads of
Sections~\ref{sec:theory} and~\ref{sec:cost} and Appendix~\ref{app:mnistcomp} were trained with
Adam (learning rate $5\times10^{-4}$, batch $128$) for $30$ epochs ($20$ for the digit $>5$
head). The ablation heads of Appendix~\ref{app:ablation} were trained with AdamW (learning rate
$3\times10^{-4}$, weight decay $10^{-4}$, batch $128$) for $30$ epochs.

\emph{Diagnostics of Section~\ref{sec:theory}.} $\hat\sigma^2(t)$ is the sample variance of
$h_\theta(X_t,t)$ over forward-noised test images, every $10$ steps; the guidance norm is the
mean $\ell_2$ norm of $\nabla\log(h_\theta+\delta)$ over forward-noised test images. For the
Monte Carlo reference (Tables~\ref{tab:field} and~\ref{tab:e1}) we take, at each $t$, $50$ test
images, $25$ of them of digit $3$, noise each once, and run $K=1{,}000$ reverse trajectories of
the frozen model from each $x_t$ in bf16 precision. Endpoints are judged by the classifier;
$h_{\mathrm{MC}}=(n_{\mathrm{sat}}+0.5)/(K+1)$, and the reference field is
\eqref{eq:tweedie} with both conditional means taken over the same trajectories. The reference
costs $50Kt$ denoiser evaluations at level $t$, from $5\times10^6$ at $t=100$ to
$4.5\times10^7$ at $t=900$. The $\delta$ sweep uses
$\delta\in\{10^{-10},10^{-8},10^{-6},10^{-4},10^{-2}\}$ at $\gamma=5$ with $3$ seeds $\times$
$1{,}000$ samples per value; the small-$h$ census uses $5{,}000$ held-out samples per noise
level.

\emph{Results of Section~\ref{sec:theory}.} The mean guidance norm is $2.32$ at $t=100$ and
$0.225$ at $t=900$. The fraction of held-out states with $h_\theta<0.01$ is $89\%$ at $t=50$,
$21.6\%$ at $t=500$, $2.4\%$ at $t=600$ and $0\%$ from $t=700$ on. Over the five values of
$\delta$, satisfaction is $94.8\pm0.5\%$.

\emph{SVDD (Section~\ref{sec:cost}).} At every step each particle draws $M$ candidates from the
base model's transition (one shared denoiser call for the mean) and keeps one of them, either
with probability proportional to its value or the one of highest value. SVDD-PM values a
candidate by the classifier's probability of digit $3$ on its posterior-mean (Tweedie)
estimate, which costs one denoiser call per candidate. The SVDD-MC value network is trained by
regression on $500{,}000$ pairs $(x_t,t)$, taken every $10$ steps along $5{,}000$ unguided
trajectories of the base model and labelled by the classifier's verdict on each trajectory's
endpoint ($10.8\%$ positive). Building these targets takes $5{,}000{,}000$ per-sample denoiser
evaluations ($233.1$\,s), and the regression $391.5$\,s. Our head was retrained under the same
harness (optimizer, epochs, batch size and steps per epoch) to time it: $340.3$\,s and no denoiser
evaluations. The SVDD comparison uses $5$ seeds $\times$ $5{,}000$ samples per method.

\subsection{CIFAR-10}
\label{app:cifardetails}

\emph{Base model.} The CIFAR-10 \citep{krizhevsky2009cifar} DDPM of \citet{ho2020ddpm}
(\nolinkurl{google/ddpm-cifar10-32}), sampled with $1000$ ancestral steps.

\emph{Predicates.} On images in $[0,1]$, with luminance $Y=0.299R+0.587G+0.114B$:
\emph{red-pixel} holds if more than $5\%$ of pixels satisfy $R>G+0.15$ and $R>B+0.15$;
\emph{edge density} holds if the mean Sobel gradient magnitude of $Y$ ($3\times3$ kernels, zero
padding) exceeds $0.8453$; \emph{low saturation} holds if the mean over pixels of
$(\max-\min)/(\max+10^{-8})$ across the colour channels is below $0.1712$; \emph{upper
luminance} holds if the mean of $Y$ over rows $0$--$15$ minus its mean over rows $16$--$31$
exceeds $0.1430$. The last three thresholds are the $75$th, $25$th and $75$th percentiles of
the statistic on the first $10{,}000$ training images. Positive rates on the $50{,}000$ training
images are $26.8\%$, $24.8\%$, $25.0\%$ and $24.5\%$. For the class constraints of
Table~\ref{tab:catship}, labels are the CIFAR-10 classes.

\emph{Heads.} $h_\theta$ has $675{,}969$ parameters: $3\times3$ convolutions
$3\to64\to64$, max-pool, $64\to128\to128$, max-pool, $128\to256$ (batch normalization and ReLU
after each), global average pooling, a $64$-dimensional sinusoidal time embedding with a
linear layer, and a head $320\to256\to128\to1$ (ReLU, dropout $0.1$ after the first layer,
sigmoid output). It is trained on the full training set with AdamW (learning rate
$3\times10^{-4}$, weight decay $10^{-4}$, batch $128$) for $50$ epochs, in $435$--$448$\,s.
The classifier-guidance baseline uses the same trunk and time embedding with a two-logit head
($676{,}098$ parameters), trained by cross-entropy on the same labels, at uniformly drawn $t$,
with the same optimizer, batch size and number of epochs ($301$--$322$\,s, plus about $4$\,s
to label the training set). It guides with the standard rule
$\epsilon\leftarrow\epsilon-\gamma\sqrt{1-\bar\alpha_t}\,g$, $g=\nabla_{x_t}\log
p_\phi(y{=}1\mid x_t,t)$, or with the normalized rule, which divides $g$ by $\|g\|$ per
sample (CG-norm). A guided sample costs $0.404$\,s with either method, against $0.386$\,s
unguided.

\emph{FID references.} Each seed contributes $1{,}000$ samples. The full-distribution reference
is the CIFAR-10 test set ($10{,}000$ images). The constraint-conditional reference contains the
test images that satisfy the predicate: $2{,}704$ (red-pixel), $2{,}538$ (edge density),
$2{,}536$ (low saturation) and $2{,}416$ (upper luminance); for cat and ship, it contains the
test images that the evaluation classifier assigns to the class.

\subsection{Stable Diffusion 1.5}
\label{app:sddetails}

\emph{Base model and sampler.} Stable Diffusion~1.5 \citep{rombach2022ldm}
(\nolinkurl{runwayml/stable-diffusion-v1-5}) at $512\times512$, DDIM \citep{song2021ddim} with
$50$ steps and classifier-free guidance scale $7.5$. The head operates on the
$4\times64\times64$ latents (VAE posterior mean scaled by $0.18215$) and is queried on the
latent the U-Net sees at every step.

\emph{Predicate and data.} An image is \emph{grayscale} if its mean HSV saturation (PIL, scaled
to $[0,1]$) is below $0.10$. The same function labels the training images, center-cropped to
$512\times512$, and scores the decoded samples. Training uses a fixed subset of $30{,}000$ COCO
train2017 images \citep{lin2014coco} ($6.1\%$ positive), split into $28{,}500$ for training and
$1{,}500$ for probing.

\emph{Heads.} $h_\theta$ has $1{,}672{,}833$ parameters: $3\times3$ convolutions
$4\to64\to128\to256\to256$ with stride $2$ and a final $256\to256$ convolution, each followed
by group normalization ($8$ groups) and SiLU, global average pooling, a $64$-dimensional time
embedding, and the head of the CIFAR-10 network. It is trained on Stable Diffusion's
$1000$-step noise schedule with AdamW (learning rate $3\times10^{-4}$, weight decay $10^{-4}$,
batch $256$) for $100$ epochs, in $313$\,s. The classifier-guidance baseline has the same trunk
with a two-logit head ($1{,}672{,}962$ parameters), trained by cross-entropy with the same
optimizer, batch size and number of epochs ($149$\,s). It guides with the normalized rule in
Table~\ref{tab:sdfull}, and with both rules in Table~\ref{tab:sdstd}. On the
probe split, the held-out accuracy of $h_\theta$ (thresholded at $0.5$) at
$t=100,300,500,700,900$ is $98.4$, $98.0$, $97.5$, $96.1$ and $94.7\%$, and that of the
classifier is $98.1$, $97.6$, $97.8$, $95.1$ and $94.5\%$.

\emph{Prompts and metrics.} The prompts are the first captions of $200$ randomly chosen COCO
train2017 images. For each seed they are shuffled with the seed and cycled to $500$ images, so
all methods see the same prompts in the same order. The constraint-conditional FID reference is
the set of $1{,}842$ grayscale images in the $30{,}000$-image subset; the full-distribution
reference is $5{,}000$ random COCO val2017 images, center-cropped to $512\times512$.
Table~\ref{tab:sdstd} uses a second prompt set, the first captions of $200$ COCO train2017
images chosen with seed $0$, with $1$ seed $\times$ $500$ images per run and the same
full-distribution reference.

\emph{SVDD-PM.} At every step, $M=5$ stochastic DDIM candidates ($\eta=1$) are drawn; each is
valued by the predicate on its decoded posterior-mean estimate (from the text-conditional
prediction), and the candidate of highest value is kept; $1$ seed $\times$ $300$ images.

\subsection{QM9}
\label{app:qm9details}

\emph{Base model and samplers.} The released GDSS checkpoint for QM9 \citep{jo2022gdss}: a
variance-preserving SDE ($\beta$ from $0.1$ to $1.0$) on node features and a
variance-exploding SDE ($\sigma$ from $0.2$ to $1.0$) on the adjacency, with $1{,}000$ steps
and node counts drawn from the training graphs. All runs integrate the reverse SDE with the
reverse-diffusion (Euler--Maruyama) predictor and the same score networks; guided runs add
$\gamma\nabla\log(h_\theta+\delta)$ to the node-feature score, and the unguided baseline is the
same sampler with $\gamma=0$. For reference, GDSS's predictor--corrector sampler
(reverse-diffusion predictor, one Langevin corrector step, signal-to-noise ratio $0.2$, scale
$0.9$) gives $66.9\%$ on the Lipinski-type filter. Each seed generates $1{,}000$
molecules in batches of $50$.

\emph{Data and predicates.} The QM9 \citep{ramakrishnan2014qm9} training split as preprocessed
by GDSS: $120{,}803$ kekulized molecules with at most nine heavy atoms, all of which decode to
valid molecules; the remaining $13{,}082$ molecules form the test split. Predicates are computed with RDKit on the
decoded SMILES, identically for training labels and generated molecules. \emph{Ring presence}
holds if the molecule has at least one ring ($89.5\%$ of training molecules). The
\emph{Lipinski-type} filter \citep{lipinski1997} requires molecular weight below $125$, at most
one hydrogen-bond donor and at most two acceptors ($35.3\%$); the thresholds were chosen by grid
search for the positive rate closest to $35\%$, since QM9 molecules are far smaller than the
drug-like molecules for which the original cutoffs were designed.

\emph{Heads.} $h_\theta$ has $320{,}321$ parameters: the node features and the adjacency rows
are each encoded by two linear layers of width $256$ with ReLU and averaged over nodes, a
$64$-dimensional time embedding is appended, and a head $576\to256\to128\to1$ with sigmoid output
follows. Training noises both the node features and the adjacency with their forward kernels and
uses Adam (learning rate $3\times10^{-4}$, batch $32$) for $150$ epochs.

\emph{SVDD-PM.} In the same Euler--Maruyama loop, each step draws $M=5$ candidate transitions,
decodes the posterior-mean estimate of each into a molecule, scores it with the predicate, and
keeps the candidate of highest value; $1$ seed $\times$ $500$ molecules, timed on a T4 GPU, like
the unguided reference it is compared with, GDSS's predictor--corrector sampler ($8.24$ against
$1.12$\,s per molecule). The unguided Euler--Maruyama baseline took $0.60$\,s per molecule on an
NVIDIA A100.

\section{Additional results}
\label{app:results}

\subsection{Tables and algorithm referenced from the main text}
\label{app:moved}

\begin{table}[H]
\caption{The learned guidance field on MNIST digit $3$. Mean cosine between
$\nabla\log h_\theta$ and the rollout reference~\eqref{eq:tweedie}, over the eligible
states among $50$ per noise level ($1{,}000$ trajectories per state; Table~\ref{tab:e1}),
and the fraction of $5{,}000$ held-out forward-noised samples with $h_\theta<0.01$.}
\label{tab:field}
\begin{center}
\begin{tabular}{lccc}
\toprule
Noise level $t$ & 500 & 700 & 900\\
\midrule
Cosine to reference & 0.50 & 0.54 & 0.50\\
$h_\theta<0.01$ (\%) & 21.6 & 0 & 0\\
\bottomrule
\end{tabular}
\end{center}
\end{table}

\begin{algorithm}[H]
\caption{Training and guided sampling}
\label{alg:main}
\begin{algorithmic}
\STATE \textbf{Train:} repeat: draw $x_0$ with label $y=\mathbf{1}_C(x_0)$, draw $t$;
\STATE \quad form $x_t, x'_{t+1}$ by independent forward noising of $x_0$;
\STATE \quad take a gradient step on
  $(h_\theta(x_t,t)-y)^2+(h_\theta(x_0,0)-y)^2+\lambda_{\mathrm{fc}}(h_\theta(x_t,t)-\mathrm{sg}[h_\theta(x'_{t+1},t+1)])^2$
\STATE \textbf{Sample:} $x_T\sim\mathcal{N}(0,I)$
\FOR{$t=T,\dots,1$}
  \STATE $g_t \leftarrow \nabla_{x_t}\log(h_\theta(x_t,t)+\delta)$
  \STATE $\tilde\epsilon \leftarrow \epsilon_\theta(x_t,t)-\gamma\sqrt{1-\bar\alpha_t}\,g_t$
  \STATE $x_{t-1}\leftarrow$ reverse step of the base sampler with $\tilde\epsilon$
\ENDFOR
\STATE \textbf{return} $x_0$
\end{algorithmic}
\end{algorithm}

\begin{table}[H]
\caption{Estimators of $h$ by training target and constraint type: SVDD
\citep{li2024svdd}, Value Matching \citep{jensen2026value}, \citet{guo2026hardconstraint}, DOIT \citep{zhu2026doit}, Doob's matching
\citep{chang2026doobmatching}, StitchVM \citep{go2026stitchvm}, classifier guidance
(CG) \citep{dhariwal2021diffusion}.}
\label{tab:grid}
\begin{center}
\begin{tabular}{lcc}
\toprule
 & Soft or bounded-below weights & Hard indicator $\mathbf{1}_C$\\
\midrule
Simulation (model law) & SVDD, Value Matching & Guo et al., DOIT\\
Forward noising (data law) & Doob's matching, StitchVM & CG; \textbf{this work}\\
\bottomrule
\end{tabular}
\end{center}
\end{table}

\subsection{The learned guidance field}
\label{app:field}

Table~\ref{tab:e1} extends Table~\ref{tab:field} to seven noise levels. From $t=400$ on, where
most states have enough satisfying endpoints to measure the reference, the cosine lies between
$0.47$ and $0.54$, and the error of $h_\theta$ against $h_{\mathrm{MC}}$ falls from $0.17$ to
$0.03$. At $t\le300$ the other half of the states have fewer than $10$ satisfying endpoints, so the eligible
states are those with $h_{\mathrm{MC}}$ near $1$. There
$\nabla\log h_t$ is small and its estimate in \eqref{eq:tweedie} rests on the few endpoints
outside $C$, so we do not interpret the cosine at these levels.

\begin{table}[H]
\caption{The learned guidance field on MNIST digit $3$ against the Monte Carlo reference of
Appendix~\ref{app:mnist} (full version of Table~\ref{tab:field}). Eligible states have at least
$10$ satisfying endpoints out of $K=1{,}000$; the cosine is averaged over eligible states, and the
root-mean-square error of $h_\theta$ against $h_{\mathrm{MC}}$ and the mean of $h_{\mathrm{MC}}$
are computed over all $50$ states.}
\label{tab:e1}
\begin{center}
\begin{tabular}{rcccc}
\toprule
$t$ & Eligible states & Mean $h_{\mathrm{MC}}$ & RMSE of $h_\theta$ & Cosine to reference\\
\midrule
100 & 25/50 & 0.500 & 0.190 & $-0.145$\\
200 & 25/50 & 0.498 & 0.161 & $-0.067$\\
300 & 25/50 & 0.456 & 0.141 & +0.240\\
400 & 28/50 & 0.356 & 0.173 & +0.467\\
500 & 42/50 & 0.234 & 0.170 & +0.501\\
700 & 50/50 & 0.122 & 0.048 & +0.544\\
900 & 50/50 & 0.111 & 0.031 & +0.499\\
\bottomrule
\end{tabular}
\end{center}
\end{table}

\subsection{CIFAR-10}
\label{app:cifar}

Table~\ref{tab:cifarfull} gives the complete sweep of ours and of the normalized rule behind
Table~\ref{tab:cifar}. At every guidance scale and under both protocols our samples are closer
to the reference than those of normalized classifier guidance: the ratio of its FID to ours
ranges from $1.39$ to $3.21$ for FID$_{\mathrm{c}}$ and from $1.39$ to $2.55$ for
FID$_{\mathrm{f}}$. Table~\ref{tab:cgstd} gives the standard rule of Table~\ref{tab:cifar}
with FID$_{\mathrm{f}}$. On the class constraints of
Table~\ref{tab:catship}, satisfaction rises from $14.2\%$ to $71.5\%$ (cat) and from $7.2\%$ to
$87.5\%$ (ship), and FID$_{\mathrm{c}}$ improves from $82.5$ and $140.4$ to at most $62.5$ and
$44.0$ for $\gamma\ge5$, while FID$_{\mathrm{f}}$ rises as the samples concentrate on one class.

\begin{table}[H]
\caption{Full sweep of normalized classifier guidance (CG-norm) and ours behind Table~\ref{tab:cifar} (the standard rule is in Table~\ref{tab:cgstd}), on four
CIFAR-10 predicates at every guidance scale, with the unguided model at $\gamma=0$. Mean over
$3$ seeds $\times$ $1{,}000$ samples; the seed standard deviation is at most $0.8$ for guided
satisfaction ($2.3$ for unguided) and $2.0$ for FID.}
\label{tab:cifarfull}
\begin{center}
\begin{tabular}{lrrrrrrr}
\toprule
 & & \multicolumn{2}{c}{Satisfaction (\%)} & \multicolumn{2}{c}{FID$_{\mathrm{c}}$} & \multicolumn{2}{c}{FID$_{\mathrm{f}}$}\\
Predicate & $\gamma$ & CG-norm & Ours & CG-norm & Ours & CG-norm & Ours\\
\midrule
Red-pixel & 0 & \multicolumn{2}{c}{31.3} & \multicolumn{2}{c}{54.0} & \multicolumn{2}{c}{42.5}\\
 & 2 & 100.0 & 97.2 & 107.3 & 48.8 & 128.6 & 59.0\\
 & 5 & 100.0 & 100.0 & 187.0 & 58.2 & 198.5 & 77.7\\
 & 10 & 100.0 & 100.0 & 209.4 & 75.1 & 218.4 & 98.0\\
 & 15 & 100.0 & 100.0 & 212.3 & 93.1 & 220.9 & 115.4\\
\midrule
Upper luminance & 0 & \multicolumn{2}{c}{23.1} & \multicolumn{2}{c}{69.9} & \multicolumn{2}{c}{42.5}\\
 & 2 & 99.9 & 99.5 & 61.7 & 39.2 & 99.6 & 57.2\\
 & 5 & 100.0 & 100.0 & 81.4 & 43.9 & 119.2 & 76.0\\
 & 10 & 100.0 & 100.0 & 86.1 & 50.0 & 124.2 & 86.3\\
 & 15 & 100.0 & 100.0 & 87.7 & 52.7 & 125.8 & 90.0\\
\midrule
Edge density & 0 & \multicolumn{2}{c}{16.5} & \multicolumn{2}{c}{61.4} & \multicolumn{2}{c}{42.5}\\
 & 2 & 99.7 & 99.3 & 68.6 & 44.6 & 90.7 & 52.3\\
 & 5 & 100.0 & 100.0 & 96.9 & 59.3 & 122.5 & 77.0\\
 & 10 & 100.0 & 100.0 & 106.3 & 73.3 & 133.1 & 92.9\\
 & 15 & 100.0 & 100.0 & 109.5 & 78.8 & 137.2 & 98.7\\
\midrule
Low saturation & 0 & \multicolumn{2}{c}{16.8} & \multicolumn{2}{c}{63.0} & \multicolumn{2}{c}{42.5}\\
 & 2 & 99.8 & 75.2 & 68.7 & 43.4 & 90.2 & 47.2\\
 & 5 & 100.0 & 95.9 & 101.6 & 45.9 & 120.6 & 58.9\\
 & 10 & 100.0 & 99.5 & 111.7 & 61.1 & 129.6 & 76.9\\
 & 15 & 100.0 & 99.9 & 116.8 & 73.2 & 134.9 & 88.7\\
\bottomrule
\end{tabular}
\end{center}
\end{table}

\begin{table}[H]
\caption{Classifier guidance with the standard rule on the four CIFAR-10 predicates of
Table~\ref{tab:cifar}. Satisfaction (\%), FID$_{\mathrm{c}}$ and FID$_{\mathrm{f}}$; mean over
$3$ seeds $\times$ $1{,}000$ samples; the seed standard deviation is at most $1.3$ for
FID$_{\mathrm{c}}$ and $0.2$ for satisfaction.}
\label{tab:cgstd}
\begin{center}
\begin{tabular}{lrrrrrr}
\toprule
 & \multicolumn{3}{c}{$\gamma=5$} & \multicolumn{3}{c}{$\gamma=15$}\\
Predicate & Sat. & FID$_{\mathrm{c}}$ & FID$_{\mathrm{f}}$ & Sat. & FID$_{\mathrm{c}}$ & FID$_{\mathrm{f}}$\\
\midrule
Red-pixel & 100.0 & 62.8 & 82.6 & 100.0 & 98.2 & 120.1\\
Upper luminance & 100.0 & 43.4 & 75.0 & 100.0 & 50.1 & 86.6\\
Edge density & 99.7 & 52.5 & 67.5 & 100.0 & 72.7 & 90.4\\
Low saturation & 99.5 & 48.6 & 62.4 & 100.0 & 71.2 & 85.6\\
\bottomrule
\end{tabular}
\end{center}
\end{table}

\begin{table}[H]
\caption{Our guidance on two CIFAR-10 class constraints. A separately trained classifier
judges the samples and selects the constraint-conditional FID reference from the test set.
Mean over $3$ seeds $\times$ $1{,}000$ samples; the seed standard deviation is at most $2.0$.}
\label{tab:catship}
\begin{center}
\begin{tabular}{lrrrr}
\toprule
Class & $\gamma$ & Satisfaction (\%) & FID$_{\mathrm{c}}$ & FID$_{\mathrm{f}}$\\
\midrule
Cat & 0 & 14.2 & 82.5 & 42.5\\
 & 2 & 32.7 & 66.6 & 54.5\\
 & 5 & 51.2 & 60.8 & 70.5\\
 & 10 & 66.2 & 60.0 & 86.2\\
 & 15 & 71.5 & 62.5 & 95.8\\
\midrule
Ship & 0 & 7.2 & 140.4 & 42.5\\
 & 2 & 50.2 & 52.7 & 77.3\\
 & 5 & 75.9 & 41.5 & 114.7\\
 & 10 & 84.6 & 42.6 & 132.8\\
 & 15 & 87.5 & 44.0 & 139.5\\
\bottomrule
\end{tabular}
\end{center}
\end{table}

\subsection{Allocation, estimand and rejection experiments}
\label{app:alloc}

These runs use the base model and heads of Appendix~\ref{app:cifardetails}, $1{,}000$ ancestral
steps unless stated, $1{,}000$ samples per run, and an NVIDIA A100; the classifier-guidance
arms use the matched classifiers of Section~\ref{sec:cifar} with the standard rule (and, in
Table~\ref{tab:alloc}, two further classifiers trained with the same recipe). Unguided pass rates
differ slightly between the sample sets of Tables~\ref{tab:cifar}, \ref{tab:gap} and the
rejection runs below ($30.8$ to $31.5\%$ on red-pixel, $21.8$ to $23.1\%$ on upper luminance),
as expected at these sample sizes.

\emph{Step profile.} Along $250$ standard-rule trajectories at $\gamma=2$, $\bar m(t)$ is the
mean over samples of $\sqrt{1-\bar\alpha_t}\,\|g\|$ (Table~\ref{tab:profile}). The rule with
an equal step gives every sample $\gamma\,\bar m(t)\,g/\|g\|$; the $1/\hat h$ rule gives sample
$i$ the step $\gamma\,\bar m(t)\,w_i\,g_i/\|g_i\|$ with $w_i=(1/\hat h_i)/\mathrm{mean}_j(1/\hat h_j)$
over the batch of $250$ and $\hat h_i=p_\phi(y{=}1\mid x_t)$ floored at $10^{-3}$. Classifiers 2
and 3 of Table~\ref{tab:alloc} are trained with training seeds $1$ and $2$ and the budget of
Appendix~\ref{app:cifardetails}; for each, $\bar m(t)$ is measured along its own $1{,}000$
standard-rule trajectories (sampling seed $0$).

\begin{table}[H]
\caption{Mean step per unit $\gamma$, $\sqrt{1-\bar\alpha_t}\,\|\nabla\log\hat h\|$ averaged over
$250$ trajectories at $\gamma=2$: ``ours'' rows along our guided trajectories, CG rows along
standard-rule trajectories of the matched classifier. The normalized rule's step is
$\sqrt{1-\bar\alpha_t}$.}
\label{tab:profile}
\begin{center}
\small
\begin{tabular}{lrrrrrrrrrr}
\toprule
$t$ & 999 & 899 & 799 & 699 & 599 & 499 & 399 & 299 & 199 & 99\\
\midrule
Red-pixel, ours & 0.067 & 0.066 & 0.118 & 0.209 & 0.276 & 0.350 & 0.376 & 0.301 & 0.232 & 0.143\\
Red-pixel, CG & 0.091 & 0.084 & 0.147 & 0.245 & 0.356 & 0.401 & 0.317 & 0.232 & 0.131 & 0.069\\
Upper lum., ours & 0.163 & 0.181 & 0.358 & 0.489 & 0.372 & 0.186 & 0.090 & 0.044 & 0.023 & 0.012\\
Upper lum., CG & 0.141 & 0.205 & 0.432 & 0.526 & 0.266 & 0.099 & 0.046 & 0.018 & 0.012 & 0.006\\
Normalized & 1.000 & 1.000 & 0.999 & 0.997 & 0.987 & 0.960 & 0.897 & 0.777 & 0.584 & 0.321\\
\bottomrule
\end{tabular}
\end{center}
\end{table}

\emph{Step and $\hat h$ across samples.} Along $250$ standard-rule trajectories at $\gamma=2$
(sampling seed $600$), Table~\ref{tab:spearman} gives the Spearman correlation between each
sample's step and $\hat h(x_t)$; every correlation has $p<10^{-24}$.

\begin{table}[H]
\caption{Spearman correlation between the per-sample step $\sqrt{1-\bar\alpha_t}\,\|g_i\|$ and
$\hat h_i=p_\phi(y{=}1\mid x_t)$ across $250$ standard-rule trajectories at $\gamma=2$.}
\label{tab:spearman}
\begin{center}
\small
\begin{tabular}{lrrrr}
\toprule
$t$ & 799 & 599 & 399 & 199\\
\midrule
Red-pixel & $-0.688$ & $-0.598$ & $-0.969$ & $-0.998$\\
Upper luminance & $-0.791$ & $-0.993$ & $-0.998$ & $-0.969$\\
\bottomrule
\end{tabular}
\end{center}
\end{table}

\emph{Estimands.} Calibration is the expected calibration error with ten equal-width bins,
averaged over ten noise levels spaced evenly along the sampler's schedule; model-law scores use
$2{,}000$ unguided trajectories (seed $300$) and their outcomes, data-law scores the $10{,}000$
forward-noised CIFAR-10 test images. With $100$ steps the sampler uses the base model's
scheduler with $100$ strided timesteps (Table~\ref{tab:gap}). Guided runs with $100$ steps
(seed $0$), satisfaction (\%) and FID$_{\mathrm{c}}$: red-pixel unguided $45.7$ / $118.1$, ours
$\gamma=2$ $99.5$ / $103.9$, CG $\gamma=2$ $99.7$ / $107.6$; upper luminance unguided $22.3$ /
$140.1$, ours $\gamma=2$ $99.5$ / $104.3$, CG $\gamma=2$ $99.6$ / $104.3$.

\begin{table}[H]
\caption{Calibration of $h_\theta$ under the model's law and the data law (mean expected
calibration error over ten noise levels), with the model's unguided pass rate over the $2{,}000$
trajectories.}
\label{tab:gap}
\begin{center}
\small
\begin{tabular}{lrrrrrr}
\toprule
 & \multicolumn{3}{c}{$1000$ steps} & \multicolumn{3}{c}{$100$ steps}\\
Predicate & Pass (\%) & Model law & Data law & Pass (\%) & Model law & Data law\\
\midrule
Red-pixel & 31.5 & 0.051 & 0.051 & 46.0 & 0.155 & 0.049\\
Upper luminance & 21.8 & 0.021 & 0.024 & 21.8 & 0.023 & 0.024\\
\bottomrule
\end{tabular}
\end{center}
\end{table}

\emph{Rejection sampling and the rare constraint.} Rejection runs draw $4{,}000$ (red-pixel)
and $4{,}500$ (upper luminance) unguided samples with seed $100$ and keep the first $1{,}000$
that pass: $1{,}230$ of $4{,}000$ ($30.8\%$) and $1{,}014$ of $4{,}500$ ($22.5\%$)
pass, and the first $1{,}000$ accepted samples have FID$_{\mathrm{c}}$ $48.4$ and $42.4$ and
FID$_{\mathrm{f}}$ $58.2$ and $51.6$. The rare constraint holds if the red-pixel fraction of the image exceeds $0.6104$,
the 99th percentile over the training set: $498$ of $50{,}000$ training and $87$ of $10{,}000$
test images pass, and its FID$_{\mathrm{c}}$ reference is all $585$. Both estimators are
retrained on these labels with the budget of Appendix~\ref{app:cifardetails}, with two training
seeds (Table~\ref{tab:rare}). Unguided and guided samples take $0.326$ and $0.345$\,s.

\begin{table}[H]
\caption{The rare constraint ($1.00\%$ of training images; the model passes $1.7\%$ of
$2{,}000$ unguided samples). Satisfaction (\%) and FID$_{\mathrm{c}}$, sampling seed $0$,
$1{,}000$ samples per run, for heads trained with seed $0$ and seed $1$; CG uses the standard
rule; --: not run.}
\label{tab:rare}
\begin{center}
\small
\begin{tabular}{lrrrrr}
\toprule
Method & $\gamma$ & Sat. (seed 0) & FID$_{\mathrm{c}}$ (seed 0) & Sat. (seed 1) & FID$_{\mathrm{c}}$ (seed 1)\\
\midrule
Ours & 2 & 99.0 & 80.2 & 95.5 & 79.8\\
 & 5 & 99.9 & 104.9 & -- & --\\
 & 10 & 100.0 & 131.9 & -- & --\\
CG & 2 & 95.4 & 78.6 & 96.0 & 80.8\\
 & 5 & 100.0 & 109.0 & -- & --\\
 & 10 & 100.0 & 149.6 & -- & --\\
\bottomrule
\end{tabular}
\end{center}
\end{table}

\subsection{Classifier accuracy across noise levels}
\label{app:diag}

\begin{figure}[H]
\centering
\includegraphics[width=\linewidth]{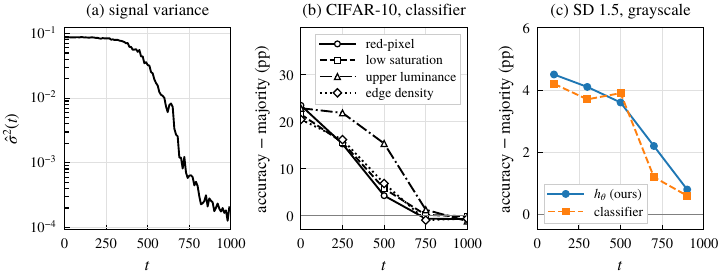}
\caption{\textbf{Signal contraction in three settings.} (a) The
constraint-signal variance $\hat\sigma^2(t)$ on MNIST falls $492\times$ between
$t=100$ and $t=900$ (Proposition~\ref{prop:contraction}). (b) On CIFAR-10, the
held-out accuracy of the matched noisy classifier above its majority baseline decays
to within 1.3 points of that baseline by $t=750$ for all four predicates. (c) On Stable Diffusion~1.5,
the held-out accuracies of $h_\theta$ and the matched classifier are within $1.0$ point of each other at every probed noise level.}
\label{fig:diagnostic}
\end{figure}

By $t=750$ every classifier is within $1.3$ points of its majority baseline
(Table~\ref{tab:cgacc}, Figure~\ref{fig:diagnostic}b). On red-pixel, our head loses its accuracy at the same rate as the
classifier. Both estimate the same function (Section~\ref{sec:estimand}); what differs between
the two methods is the drift applied once the signal is gone.

\begin{table}[H]
\caption{Held-out accuracy (\%) across noise levels. Classifier rows: the matched noisy-state
classifiers of Section~\ref{sec:cifar} on the first $2{,}000$ CIFAR-10 test images
(Figure~\ref{fig:diagnostic}b). $h_\theta$ row: our head for red-pixel, thresholded at $0.5$, on
the $4{,}000$ test images of Figure~\ref{fig:cifar_h}, of which $27.1\%$ are positive. The
majority baseline is $100$ minus the training positive rate.}
\label{tab:cgacc}
\begin{center}
\begin{tabular}{llrrrrrr}
\toprule
Predicate & Estimator & $t=0$ & $250$ & $500$ & $750$ & $999$ & Majority\\
\midrule
Red-pixel & Classifier & 96.7 & 88.5 & 77.5 & 72.6 & 72.5 & 73.2\\
 & $h_\theta$ & 94.3 & 87.2 & 77.6 & 72.8 & 72.9 & 73.2\\
Upper luminance & Classifier & 98.4 & 97.4 & 90.9 & 76.8 & 74.4 & 75.5\\
Edge density & Classifier & 95.7 & 91.4 & 82.1 & 74.3 & 74.6 & 75.2\\
Low saturation & Classifier & 96.6 & 90.6 & 80.8 & 75.2 & 74.9 & 75.0\\
\bottomrule
\end{tabular}
\end{center}
\end{table}

\subsection{Comparison with SVDD on MNIST}
\label{app:svdd}

With proportional selection SVDD-PM needs $M=20$ candidates, at $21\times$ our time per sample,
to match our satisfaction; keeping the best of $M=5$ candidates reaches $100.0\%$ at $5.7\times$
our time per sample (Table~\ref{tab:svdd}). Every SVDD-PM step also evaluates the checker on the
posterior-mean estimates of $M$ candidates, whereas our sampler never calls it.

\begin{table}[H]
\caption{Guidance on MNIST digit $3$ against SVDD-PM, $5$ seeds $\times$ $5{,}000$ samples,
our head and the SVDD-MC value network have $122{,}657$ parameters each. Time is per sample on one RTX~4090.
Training the SVDD-MC value network took $624.6$\,s, of which $233.1$\,s built the targets from
$5{,}000{,}000$ denoiser evaluations; training our head took $340.3$\,s and no denoiser
evaluations.}
\label{tab:svdd}
\begin{center}
\begin{tabular}{lrlrr}
\toprule
Method & $M$ & Candidate selection & Satisfaction (\%) & s/sample\\
\midrule
Ours ($\gamma=5$) & -- & -- & 95.1 $\pm$ 0.3 & 0.050\\
SVDD-PM & 5 & proportional to value & 92.5 $\pm$ 0.3 & 0.294\\
SVDD-PM & 20 & proportional to value & 95.2 $\pm$ 0.3 & 1.048\\
SVDD-PM & 5 & highest value & 100.0 $\pm$ 0.0 & 0.283\\
\bottomrule
\end{tabular}
\end{center}
\end{table}

\subsection{Objective ablations}
\label{app:ablation}

The method needs only $\mathcal L_{\mathrm{sup}}$ (Table~\ref{tab:e10}): with three training
seeds per objective, $\mathcal L_{\mathrm{sup}}$ alone reaches $86.7\pm3.3\%$, and removing
either regularizer, or both, does not lower satisfaction. The four objectives reach $81.3$ to
$86.7\%$, and the spread across training seeds (standard deviation
$1.5$ to $3.3$ points) is several times larger than the spread across sampling seeds
($0.14$ to $0.44$ points). This is what Proposition~\ref{prop:target} predicts: $\mathcal
L_{\mathrm{sup}}$ alone identifies $h$. The heads of Section~\ref{sec:experiments} were trained with the full
objective before this ablation and are reported unchanged; we attribute no result to the
regularizers.
Among constructions of the consistency pair (Table~\ref{tab:e7}), independent noisings of the same
$x_0$, the construction of Section~\ref{sec:method}, are best at $\gamma=2$ and within $0.4$
points of the best at $\gamma\ge5$.

\begin{table}[H]
\caption{Loss ablation on MNIST digit $3$ at $\gamma=3$ (strict fp32). Three heads per objective
(training seeds $0$--$2$), each evaluated over $5$ sampling seeds $\times$ $5{,}000$ samples;
mean and standard deviation across training seeds. For every head, the standard deviation across
sampling seeds is $0.14$ to $0.44$ points.}
\label{tab:e10}
\begin{center}
\begin{tabular}{lcc}
\toprule
Objective & Per training seed (\%) & Mean (\%)\\
\midrule
$\mathcal L_{\mathrm{bdry}}+\mathcal L_{\mathrm{sup}}+0.1\,\mathcal L_{\mathrm{fc}}$ (full) & 79.2 / 79.9 / 84.9 & 81.3 $\pm$ 2.5\\
$\mathcal L_{\mathrm{bdry}}+\mathcal L_{\mathrm{sup}}$ & 85.3 / 83.9 / 89.6 & 86.3 $\pm$ 2.4\\
$\mathcal L_{\mathrm{sup}}+0.1\,\mathcal L_{\mathrm{fc}}$ & 87.4 / 83.9 / 86.7 & 86.0 $\pm$ 1.5\\
$\mathcal L_{\mathrm{sup}}$ only & 88.2 / 82.2 / 89.8 & 86.7 $\pm$ 3.3\\
\bottomrule
\end{tabular}
\end{center}
\end{table}

\begin{table}[H]
\caption{Construction of the consistency pair in $\mathcal L_{\mathrm{fc}}$ on MNIST digit $3$:
one head per construction (training seed $0$, default GPU precision), evaluated over $5$ seeds
$\times$ $5{,}000$ samples. Satisfaction (\%), mean $\pm$ standard deviation over seeds.}
\label{tab:e7}
\begin{center}
\begin{tabular}{lccc}
\toprule
Pair $(x_t,x'_{t+1})$ & $\gamma=2$ & $\gamma=5$ & $\gamma=10$\\
\midrule
Independent noisings of $x_0$ (Section~\ref{sec:method}) & 76.7 $\pm$ 0.6 & 94.0 $\pm$ 0.1 & 98.3 $\pm$ 0.1\\
Shared noise draw & 72.8 $\pm$ 0.6 & 94.1 $\pm$ 0.2 & 98.7 $\pm$ 0.1\\
$x'_{t+1}$ drawn from $q(x_{t+1}\mid x_t)$ & 63.2 $\pm$ 0.5 & 87.8 $\pm$ 0.3 & 96.1 $\pm$ 0.2\\
\bottomrule
\end{tabular}
\end{center}
\end{table}

\subsection{Composition on MNIST}
\label{app:mnistcomp}

Guiding with one head raises its own constraint but leaves the joint rate between $25.7\%$ and
$31.8\%$. Adding the two log-gradients raises the joint rate from $16.1\%$ to $67.3\%$ at
$\gamma=10$ with equal weights, and to $74.3\%$ when the even head is weighted twice; the weights
move satisfaction between the two constraints (Table~\ref{tab:mnistcomp}).

\begin{table}[H]
\caption{Composition of two separately trained MNIST heads, digit even and digit $>5$, whose
intersection is $\{6,8\}$. Entries: joint satisfaction (\%), followed by the two marginal rates
(even, $>5$) in parentheses; mean over $5$ seeds $\times$ $5{,}000$ samples (seed standard
deviation at most $1.0$). Weights $(w_1,w_2)$ multiply $\log h_{\mathrm{even}}$ and
$\log h_{>5}$; $(1,0.5)$ at $\gamma=10$ applies the same guidance as $(2,1)$ at $\gamma=5$, as does
$(0.5,1)$ at $\gamma=10$ with $(1,2)$ at $\gamma=5$.}
\label{tab:mnistcomp}
\begin{center}
\begin{tabular}{lccc}
\toprule
Guidance & $\gamma=2$ & $\gamma=5$ & $\gamma=10$\\
\midrule
Unguided & \multicolumn{3}{c}{16.1 \;(53.7, 33.3)}\\
Even only & 25.7 \;(84.9, 31.0) & 26.9 \;(92.5, 29.6) & 26.8 \;(95.7, 28.5)\\
$>5$ only & 29.5 \;(51.4, 63.2) & 31.8 \;(45.7, 76.8) & 29.9 \;(39.4, 84.6)\\
Composed, $(1,1)$ & 43.2 \;(81.6, 56.5) & 56.0 \;(87.1, 66.7) & 67.3 \;(91.3, 74.8)\\
Composed, $(2,1)$ & 46.4 \;(88.1, 54.9) & 60.4 \;(92.1, 66.9) & 74.3 \;(95.5, 78.2)\\
Composed, $(1,2)$ & 48.0 \;(78.3, 65.7) & 61.0 \;(85.2, 74.1) & 71.4 \;(90.3, 80.1)\\
Composed, $(0.5,1)$ & 39.1 \;(72.5, 58.9) & 50.9 \;(79.8, 67.7) & 60.9 \;(85.2, 74.0)\\
Composed, $(1,0.5)$ & 37.3 \;(83.8, 46.9) & 49.6 \;(89.1, 57.8) & 60.3 \;(92.0, 66.9)\\
\bottomrule
\end{tabular}
\end{center}
\end{table}

\subsection{Stable Diffusion 1.5}
\label{app:sd}

\begin{table}[H]
\caption{Stable Diffusion~1.5, grayscale predicate, 3 seeds $\times$ 500 images
(SVDD-PM: 1 seed $\times$ 300, $M=5$; Wilson 95\% interval 6.3--12.8\%). CLIP score of \citet{radford2021clip};
FID$_{\mathrm{c}}$ and FID$_{\mathrm{f}}$ against the references of Appendix~\ref{app:sddetails}.}
\label{tab:sd}
\begin{center}
\begin{tabular}{lrrrrrr}
\toprule
Method & $\gamma$ / $M$ & Sat. (\%) & CLIP & FID$_{\mathrm{c}}$ & FID$_{\mathrm{f}}$ & s/sample\\
\midrule
Unguided & 0 & 7.2 & 26.19 & 109.9 & 88.9 & 0.81\\
Ours & 10 & 33.1 & 26.23 & 102.6 & 89.9 & 1.03\\
Ours & 35 & 51.1 & 26.08 & 99.6 & 92.0 & 1.03\\
Ours & 75 & 60.0 & 25.96 & 98.2 & 93.5 & 1.03\\
SVDD-PM & $M=5$ & 9.0 & 26.47 & 133.9 & 117.6 & 16.65\\
\bottomrule
\end{tabular}
\end{center}
\end{table}

Our satisfaction increases with $\gamma$ at every step of the sweep, from $33.1\%$ at
$\gamma=10$ to $60.0\%$ at $\gamma=75$, while the CLIP score stays within $0.23$ of the unguided
value and FID$_{\mathrm{c}}$ improves from $109.9$ to between $98.2$ and $102.6$. Normalized
classifier guidance is highest at $\gamma=10$ ($65.1\%$): it rises from $12.3\%$ at $\gamma=1$
and declines as $\gamma$ grows, to $24.7\%$ at $\gamma=75$ (Table~\ref{tab:sdfull}).

\begin{table}[H]
\caption{Full sweep behind Table~\ref{tab:sd}: grayscale predicate on Stable Diffusion~1.5,
$3$ seeds $\times$ $500$ images; CG-norm is the matched classifier with the normalized
rule. Satisfaction is the mean $\pm$ standard deviation over seeds; CLIP score and FID are means.
Rows marked $\dagger$ were run in a separate session with the same prompts,
references and seeds.}
\label{tab:sdfull}
\begin{center}
\begin{tabular}{lrrrrr}
\toprule
Method & $\gamma$ & Satisfaction (\%) & CLIP & FID$_{\mathrm{c}}$ & FID$_{\mathrm{f}}$\\
\midrule
Unguided & 0 & 7.2 $\pm$ 1.1 & 26.19 & 109.9 & 88.9\\
Ours & 10 & 33.1 $\pm$ 1.1 & 26.23 & 102.6 & 89.9\\
Ours & 20 & 44.2 $\pm$ 0.6 & 26.11 & 100.4 & 91.0\\
Ours & 35 & 51.1 $\pm$ 1.4 & 26.08 & 99.6 & 92.0\\
Ours & 50 & 54.5 $\pm$ 1.4 & 26.01 & 100.0 & 93.6\\
Ours & 75 & 60.0 $\pm$ 0.5 & 25.96 & 98.2 & 93.5\\
CG-norm$^\dagger$ & 1 & 12.3 $\pm$ 0.7 & 26.21 & 109.2 & 89.4\\
CG-norm$^\dagger$ & 2.5 & 20.3 $\pm$ 1.5 & 26.22 & 106.9 & 89.8\\
CG-norm$^\dagger$ & 5 & 43.4 $\pm$ 2.0 & 26.20 & 103.7 & 90.8\\
CG-norm$^\dagger$ & 7.5 & 61.6 $\pm$ 3.6 & 26.10 & 100.4 & 91.8\\
CG-norm & 10 & 65.1 $\pm$ 0.9 & 26.09 & 99.3 & 92.5\\
CG-norm & 20 & 56.2 $\pm$ 1.6 & 26.00 & 101.1 & 92.7\\
CG-norm & 35 & 43.1 $\pm$ 2.9 & 25.96 & 101.1 & 92.3\\
CG-norm & 50 & 34.1 $\pm$ 2.6 & 25.87 & 101.1 & 92.0\\
CG-norm & 75 & 24.7 $\pm$ 2.3 & 25.62 & 103.5 & 94.0\\
\bottomrule
\end{tabular}
\end{center}
\end{table}

\begin{table}[H]
\caption{Stable Diffusion~1.5 on the second prompt set of Appendix~\ref{app:sddetails}
($1$ seed $\times$ $500$ images): ours, the matched classifier with the standard rule (CG) and
with the normalized rule (CG-norm). Satisfaction (\%), CLIP score and FID$_{\mathrm{f}}$.}
\label{tab:sdstd}
\begin{center}
\begin{tabular}{lrrrr}
\toprule
Method & $\gamma$ & Satisfaction (\%) & CLIP & FID$_{\mathrm{f}}$\\
\midrule
Unguided & 0 & 6.4 & 26.77 & 87.25\\
Ours & 10 & 30.0 & 26.62 & 89.65\\
Ours & 35 & 48.8 & 26.52 & 90.89\\
Ours & 75 & 58.0 & 26.37 & 92.96\\
CG & 10 & 28.8 & 26.60 & 89.72\\
CG & 35 & 46.6 & 26.50 & 92.35\\
CG & 75 & 54.4 & 26.26 & 91.92\\
CG-norm & 10 & 61.8 & 26.40 & 91.66\\
CG-norm & 75 & 25.4 & 26.06 & 93.78\\
\bottomrule
\end{tabular}
\end{center}
\end{table}

\subsection{QM9}
\label{app:qm9}

On the Lipinski-type filter satisfaction rises with $\gamma$ from $88.7\%$ to $99.5\%$ while
uniqueness falls from $92.4\%$ to $63.2\%$, and novelty stays at or above $92.8\%$, against $54.3\%$ for the unguided model with the same
sampler (Table~\ref{tab:qm9full}). With
composed guidance the joint rate is within $0.7$ points of the product of the guided marginals at
every $\gamma\ge5$ (Table~\ref{tab:qm9comp}).

\begin{table}[H]
\caption{Full QM9 results for the Lipinski-type filter, $3$ seeds $\times$ $1{,}000$ molecules, all with the Euler--Maruyama sampler (satisfaction: mean $\pm$
standard deviation over seeds). $^{\dagger}$SVDD-PM: $1$ seed $\times$ $500$ molecules on a T4
GPU (Appendix~\ref{app:qm9details}).}
\label{tab:qm9full}
\begin{center}
\begin{tabular}{llrrrr}
\toprule
Method & $\gamma$ / $M$ & Satisfaction (\%) & Valid (\%) & Unique (\%) & Novel (\%)\\
\midrule
\multicolumn{6}{l}{\emph{Lipinski-type filter}}\\
Unguided & 0 & 54.3 $\pm$ 0.9 & 100.0 & 98.0 & 98.3\\
Ours & 2 & 88.7 $\pm$ 0.4 & 100.0 & 92.4 & 96.2\\
Ours & 5 & 94.1 $\pm$ 0.4 & 100.0 & 88.8 & 95.2\\
Ours & 10 & 96.8 $\pm$ 0.2 & 100.0 & 83.2 & 94.5\\
Ours & 20 & 98.1 $\pm$ 0.2 & 100.0 & 76.0 & 94.4\\
Ours & 50 & 99.5 $\pm$ 0.4 & 100.0 & 63.2 & 92.8\\
SVDD-PM$^{\dagger}$ & $M{=}5$ & 96.0 & 100.0 & 97.8 & 97.1\\
\bottomrule
\end{tabular}
\end{center}
\end{table}

\begin{table}[H]
\caption{Composition on QM9: separately trained Lipinski-type and ring-presence heads, guided by
$\log h_{\mathrm{Lip}}+\log h_{\mathrm{ring}}$; $\gamma=0$ is the unguided
model with the same sampler, the run of Table~\ref{tab:qm9full}. Joint
satisfaction (mean $\pm$ standard deviation over $3$ seeds $\times$ $1{,}000$ molecules), the two
marginal rates, and the product of the two rates shown.}
\label{tab:qm9comp}
\begin{center}
\begin{tabular}{rrrrrrr}
\toprule
$\gamma$ & Joint (\%) & Lipinski (\%) & Ring (\%) & Product (\%) & Unique (\%) & Novel (\%)\\
\midrule
0 & 39.6 $\pm$ 0.4 & 54.3 & 76.7 & 41.6 & 98.0 & 98.3\\
5 & 72.4 $\pm$ 0.4 & 92.9 & 78.6 & 73.0 & 92.3 & 96.0\\
10 & 75.5 $\pm$ 0.6 & 95.1 & 79.8 & 75.9 & 91.5 & 95.6\\
20 & 78.3 $\pm$ 0.5 & 96.3 & 81.6 & 78.6 & 90.8 & 95.9\\
\bottomrule
\end{tabular}
\end{center}
\end{table}

\subsection{Additional figures}
\label{app:figs}

\begin{figure}[H]
\centering
\includegraphics[width=\linewidth]{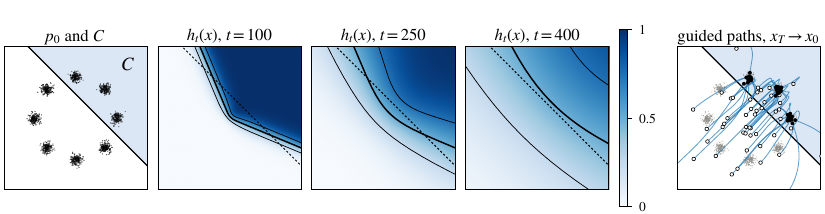}
\caption{\textbf{The constraint posterior $h_t$ on a two-dimensional example, computed exactly.}
First panel: $p_0$ is a mixture of eight Gaussians and $C$ is a half-plane (shaded), with
$\Pr(X_0\in C)=0.375$. Next three panels: $h_t(x)=\Pr(X_0\in C\mid X_t=x)$ at three noise
levels; black curves are the $0.25$, $0.5$ and $0.75$ level sets and the dashed line is the
boundary of $C$. As $t$ grows, $h_t$ flattens toward $\Pr(X_0\in C)$
(Proposition~\ref{prop:contraction}). Last panel: deterministic reverse paths guided by
$\nabla\log h_t$ with $\gamma=1$, from $x_T$ (open circles) to $x_0$ (filled). All
2,000 guided samples land in $C$, against $38.6\%$ without guidance.}
\label{fig:toy}
\end{figure}

\begin{figure}[H]
\centering
\includegraphics[width=\linewidth]{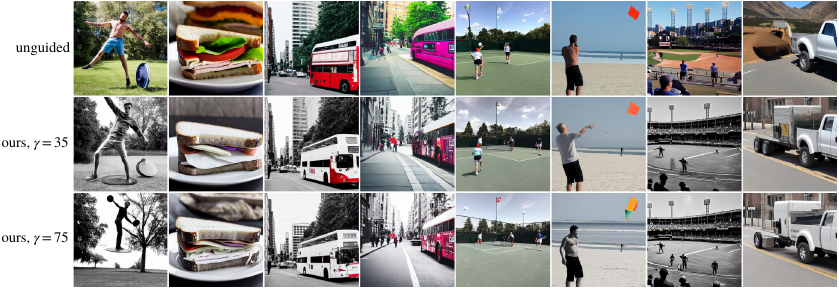}
\caption{\textbf{Stable Diffusion~1.5, grayscale constraint.} The first eight COCO captions in
our sample (columns) and, in rows, the unguided sample and ours at $\gamma=35$ and $\gamma=75$ from the same seed (DDIM, 50 steps). Guidance with $h_\theta$ reduces
colour while largely keeping the layout of the unguided image.}
\label{fig:sd_samples}
\end{figure}

\begin{figure}[H]
\centering
\includegraphics[width=0.5\linewidth]{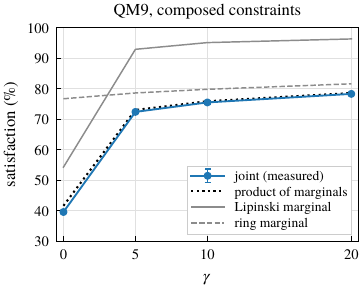}
\caption{\textbf{QM9 composition} with separately trained Lipinski-type and ring-presence
heads, guided by $\log h_{\mathrm{Lip}}+\log h_{\mathrm{ring}}$. Joint satisfaction (mean
$\pm$ seed standard deviation, 3 seeds $\times$ 1{,}000 molecules) against the product of
the two guided marginal rates, with the marginals shown for reference.}
\label{fig:composition}
\end{figure}

Figure~\ref{fig:toy} computes $h_t$ exactly on a two-dimensional example,
Figure~\ref{fig:sd_samples} shows paired Stable Diffusion samples (Section~\ref{sec:sd}),
and Figure~\ref{fig:composition} plots the QM9 composition of Table~\ref{tab:qm9comp}.

\section{Further experiments}
\label{app:further}

All runs in this section use $1{,}000$ samples per run and seed, the CIFAR-10 base model, heads,
predicates and FID references of Appendix~\ref{app:cifardetails}, and an NVIDIA A100 unless
stated (the class-constraint runs of Table~\ref{tab:allocclass} used an NVIDIA L4). The molecular
runs of Appendix~\ref{app:qm9alloc} use the QM9 model, head and metrics of
Appendix~\ref{app:qm9details} and were run partly on CPU.

\subsection{Comparison with Spectral Guidance}
\label{app:sg}

\emph{Their method.} We use the authors' released code
(\nolinkurl{github.com/gabmoreira/spectralguidance}, commit \texttt{4a1f266}). No trained network
is released, so we trained their feature network $f_\phi$ with their CIFAR-10 configuration
($K=512$ singular functions, base width $256$, batch $2048$, learning rate $10^{-4}$, ridge
$10^{-3}$, $1{,}000$ epochs) on the same base model as ours, and ran their encoding script on the
$50{,}000$ training images at $t=10,20,\dots,990$. The mean of the $16$ largest values of the
learned spectrum is $0.91$ for $t\in[300,400)$, $0.56$ for $t\in[500,600)$ and $0.05$ for
$t\ge800$, consistent with the transition near $t\approx400$ that they report. The guidance
signal is the binary label $\mathbf 1_C(x_0)$ of each training image, projected onto their basis
as in their code.

\emph{Sampler.} Both methods are run in their sampler with the settings of their example
notebook: DDIM with $100$ steps and $\eta=1$, the base network in half precision, guidance at
every step, and the update $x\leftarrow x+\kappa\sqrt{1-\bar\alpha_t}\,s$ with strength $\kappa$.
For Spectral Guidance $s$ is their released rule, $\nabla\hat p/\hat p$ rescaled per sample to a
norm in $[0.1,0.3]$. For ours $s=\nabla\log h_\theta$ with the heads of
Appendix~\ref{app:cifardetails}, unrescaled. With deterministic cuDNN, our copy of their loop
reproduces their \texttt{sample\_batch} exactly (maximum absolute difference $0$ over $16$
samples on both predicates).

\emph{Tuning and reporting.} Each method is swept over $\kappa$ on sampling seed $0$
($8$ values for ours, $9$ and $10$ for Spectral Guidance). For each
method and predicate we select the strength with the lowest FID$_{\mathrm{c}}$ among those with at
least $99.9\%$ satisfaction on seed $0$. The comparison is then reported on seeds $1$ and $2$,
which play no part in the selection (Table~\ref{tab:sg}). Satisfaction, FID$_{\mathrm{c}}$ and FID$_{\mathrm{f}}$ follow
Appendix~\ref{app:protocol} and~\ref{app:cifardetails}: the constraint-conditional reference is
the $2{,}704$ (red-pixel) or $2{,}416$ (upper luminance) test images that satisfy the predicate,
and the full reference is the $10{,}000$ test images.

\begin{table}[H]
\caption{Spectral Guidance and ours in the Spectral Guidance sampler (DDIM, $100$ steps,
$\eta=1$), each at its selected strength $\kappa$: mean over the two evaluation seeds ($1$ and
$2$) $\times$ $1{,}000$ samples; the tuning seed $0$ is excluded. Each seed is within $0.1$ of
the mean in satisfaction and within $1.3$ in FID. Time is wall-clock sampling time per sample.}
\label{tab:sg}
\begin{center}
\small
\setlength{\tabcolsep}{4pt}
\begin{tabular}{llrrrrr}
\toprule
Predicate & Method & $\kappa$ & Sat. (\%) & FID$_{\mathrm{c}}$ & FID$_{\mathrm{f}}$ & s/sample\\
\midrule
Red-pixel & Spectral Guidance & 1.5 & 99.85 & 76.07 & 99.18 & 0.039\\
 & Ours & 0.25 & 99.70 & 51.32 & 63.31 & 0.025\\
Upper luminance & Spectral Guidance & 1.0 & 99.50 & 49.41 & 85.70 & 0.039\\
 & Ours & 0.25 & 99.95 & 43.96 & 66.21 & 0.025\\
\bottomrule
\end{tabular}
\end{center}
\end{table}

On the evaluation seeds our FID$_{\mathrm{c}}$ is $24.8$ lower on red-pixel and $5.5$ lower on
upper luminance, at satisfaction within $0.5$ points. Including the tuning seed changes each mean
by at most $0.3$ (three-seed FID$_{\mathrm{c}}$: $51.10$ against $76.00$, and $43.83$ against
$49.42$).

\emph{Timing.} The time in Table~\ref{tab:sg} is the wall-clock time of the complete sampling
loop for $1{,}000$ samples in batches of $250$ on one A100, divided by $1{,}000$. It includes the
$100$ base-model evaluations and the guidance gradient at each step, and excludes image saving,
FID computation and all training. The same loop without guidance takes $0.023$\,s per sample, so
guidance adds about $0.002$\,s with our head and $0.016$\,s with $f_\phi$; the difference of
$35\%$ in total sampling time is the cost of differentiating a larger network ($9$M parameters
as reported by the authors, against $0.68$M).

\emph{Estimates of $h$.} On the $10{,}000$ test images, forward-noised with the same noise for
both methods, Table~\ref{tab:sgauroc} scores each estimate of $h_t$ against the label
$\mathbf 1_C(x_0)$ by the area under the ROC curve, which does not depend on the scale of the
estimate.

\begin{table}[H]
\caption{AUROC of the estimate of $h_t$ against $\mathbf 1_C(x_0)$ on $10{,}000$ forward-noised
test images: ours / Spectral Guidance.}
\label{tab:sgauroc}
\begin{center}
\small
\begin{tabular}{lrrrrrr}
\toprule
$t$ & 10 & 100 & 300 & 500 & 700 & 900\\
\midrule
Red-pixel & .996 / .894 & .987 / .895 & .922 / .850 & .792 / .767 & .616 / .585 & .523 / .499\\
Upper luminance & .999 / .984 & .999 / .984 & .995 / .980 & .972 / .959 & .817 / .798 & .582 / .522\\
\bottomrule
\end{tabular}
\end{center}
\end{table}

\emph{Step allocation with their estimator.} Table~\ref{tab:sgalloc} applies the test of
Section~\ref{sec:alloc} to the Spectral Guidance estimate $\hat p$: the unrescaled step
$\nabla\hat p/\hat p$, an equal step per sample at that rule's mean step $\bar m(t)$, and the
$1/\hat p$-weighted step at the same mean ($\hat p$ floored at $10^{-3}$); $\bar m(t)$ is measured
for each seed along its own unrescaled trajectories.

\begin{table}[H]
\caption{The test of Table~\ref{tab:alloc} with the Spectral Guidance estimator, in their
sampler: satisfaction (\%, mean $\pm$ standard deviation) / mean FID$_{\mathrm{c}}$ over $3$
sampling seeds $\times$ $1{,}000$ samples.}
\label{tab:sgalloc}
\begin{center}
\small
\begin{tabular}{lrlll}
\toprule
Predicate & $\kappa$ & $\nabla\hat p/\hat p$ & Equal & $1/\hat p$\\
\midrule
Red-pixel & 0.25 & 78.97 $\pm$ 1.13 / 59.0 & 73.10 $\pm$ 1.34 / 56.9 & 77.33 $\pm$ 1.52 / 58.9\\
 & 0.5 & 94.53 $\pm$ 0.57 / 66.5 & 89.27 $\pm$ 0.78 / 63.6 & 92.70 $\pm$ 0.91 / 65.7\\
Upper luminance & 0.25 & 95.60 $\pm$ 0.22 / 42.2 & 92.53 $\pm$ 2.97 / 43.2 & 96.83 $\pm$ 1.75 / 43.3\\
 & 0.5 & 99.87 $\pm$ 0.05 / 46.0 & 98.07 $\pm$ 0.65 / 45.1 & 99.87 $\pm$ 0.12 / 46.0\\
\bottomrule
\end{tabular}
\end{center}
\end{table}

The equal step has the lower mean satisfaction in all four rows ($5.9$, $5.3$, $3.1$ and $1.8$
points). The difference exceeds the sum of the two seed standard deviations in three of the four rows.
FID$_{\mathrm{c}}$ of the equal step is within $3$ of the unrescaled step in every row.

\subsection{The matched-budget test in further configurations}
\label{app:alloc2}

Table~\ref{tab:allocconfigs} lists every configuration in which the test of
Section~\ref{sec:alloc} has been run; a configuration is one estimator, constraint, guidance scale
and sampler. Tables~\ref{tab:alloc2} and~\ref{tab:allocclass} give the results beyond
Table~\ref{tab:alloc}, with the protocol of Appendix~\ref{app:alloc}; for the class constraints,
binary noisy-state classifiers are trained on the dataset labels with the recipe of
Appendix~\ref{app:cifardetails} and samples are judged as in Table~\ref{tab:catship}.

\begin{table}[H]
\caption{Configurations of the matched-budget test. ``Drop'' is the range of the loss in
satisfaction (points) of the equal step against the standard rule. The last row is on QM9; all
others are on CIFAR-10.}
\label{tab:allocconfigs}
\begin{center}
\footnotesize
\setlength{\tabcolsep}{4pt}
\begin{tabular}{>{\raggedright\arraybackslash}p{2.5cm}>{\raggedright\arraybackslash}p{1.9cm}>{\raggedright\arraybackslash}p{2.1cm}>{\raggedright\arraybackslash}p{1.6cm}>{\raggedright\arraybackslash}p{2.0cm}rr}
\toprule
Set & Estimator (training seeds) & Constraints & Scales & Sampler; sampling seeds & \# & Drop\\
\midrule
Table~\ref{tab:alloc} & classifier (0, 1, 2) & red-pixel, upper lum. & $\gamma=2$ & 1000 steps; 0 (and 1 for classifier 1) & 6 & 17.0--18.5\\
Further scales & classifier (0) & red-pixel, upper lum. & $\gamma\in\{1,5\}$ & 1000; 0 & 4 & 8.4--22.4\\
Further predicates & classifier (0) & edge density, low saturation & $\gamma=2$ & 1000; 0 & 2 & 18.5--22.2\\
Coarser samplers & classifier (0) & red-pixel, upper lum. & $\gamma=2$ & 100 and 50 steps; 0 & 4 & 6.2--17.4\\
Our head & $h_\theta$ (0) & red-pixel, upper lum. & $\gamma=2$ & 1000; 0 & 2 & 17.8--18.0\\
Class constraints & classifier (0) & cat, ship & $\gamma\in\{2,5,10\}$ & 1000; 0 (and 1 at $\gamma=5$) & 6 & 1.2--8.5\\
Spectral Guidance estimator & $f_\phi$ (one network) & red-pixel, upper lum. & $\kappa\in\{0.25,0.5\}$ & DDIM 100; 0, 1, 2 & 4 & 1.8--5.9\\
Dependent conjunction & two heads $h_\theta$, composed & red-pixel $\wedge$ low saturation & $\gamma\in\{5,10\}$ & 1000; 0, 1, 2 & 2 & 5.2--7.3\\
Molecular filter (QM9) & $h_\theta$ (0) & Lipinski-type & $\gamma\in\{2,5\}$ & 1000; 0, 1, 2 & 2 & 1.1--2.7\\
\midrule
Total & & 8 constraints & & & 32 & \\
\bottomrule
\end{tabular}
\end{center}
\end{table}

\begin{table}[H]
\caption{The test of Table~\ref{tab:alloc} at further scales, predicates and samplers, and with
our head: satisfaction (\%) / FID$_{\mathrm{c}}$, sampling seed $0$, $1{,}000$ samples.}
\label{tab:alloc2}
\begin{center}
\small
\begin{tabular}{llrrr}
\toprule
Predicate & Setting & Standard & Equal & $1/\hat h$\\
\midrule
Red-pixel & classifier, $\gamma=1$ & 85.6 / 48.7 & 67.4 / 48.3 & 74.4 / 49.0\\
 & classifier, $\gamma=5$ & 100.0 / 64.3 & 87.7 / 64.2 & 92.3 / 63.4\\
 & classifier, $\gamma=2$, 100 steps & 99.7 / 111.2 & 86.5 / 108.7 & 97.5 / 112.3\\
 & classifier, $\gamma=2$, 50 steps & 100.0 / 156.6 & 93.8 / 155.5 & 99.8 / 157.5\\
 & $h_\theta$, $\gamma=2$ & 97.8 / 50.6 & 79.8 / 50.4 & 93.0 / 50.3\\
\midrule
Upper luminance & classifier, $\gamma=1$ & 93.8 / 42.2 & 71.4 / 44.9 & 87.8 / 43.5\\
 & classifier, $\gamma=5$ & 100.0 / 43.4 & 91.6 / 43.2 & 96.1 / 43.3\\
 & classifier, $\gamma=2$, 100 steps & 99.6 / 102.5 & 82.2 / 108.2 & 91.2 / 103.6\\
 & classifier, $\gamma=2$, 50 steps & 99.9 / 144.2 & 82.7 / 147.7 & 93.3 / 145.0\\
 & $h_\theta$, $\gamma=2$ & 99.3 / 39.4 & 81.5 / 40.3 & 95.0 / 39.6\\
\midrule
Edge density & classifier, $\gamma=2$ & 93.9 / 44.9 & 71.7 / 45.3 & 84.9 / 45.1\\
Low saturation & classifier, $\gamma=2$ & 92.0 / 44.0 & 73.5 / 43.9 & 83.8 / 43.7\\
\bottomrule
\end{tabular}
\end{center}
\end{table}

\begin{table}[H]
\caption{The test of Table~\ref{tab:alloc} on two class constraints: satisfaction (\%) /
FID$_{\mathrm{c}}$, $1{,}000$ samples per run. Unguided: cat $16.6\%$, ship $7.3\%$.}
\label{tab:allocclass}
\begin{center}
\small
\begin{tabular}{lrrrrr}
\toprule
Class & $\gamma$ & Sampling seed & Standard & Equal & $1/\hat h$\\
\midrule
Cat & 2 & 0 & 42.8 / 62.2 & 41.6 / 61.2 & 42.0 / 62.7\\
 & 5 & 0 & 64.3 / 58.6 & 60.3 / 58.1 & 63.0 / 58.8\\
 & 5 & 1 & 68.0 / 60.0 & 65.9 / 58.7 & 67.1 / 59.9\\
 & 10 & 0 & 76.4 / 63.2 & 72.0 / 61.6 & 75.5 / 62.8\\
\midrule
Ship & 2 & 0 & 54.9 / 51.2 & 51.0 / 55.6 & 51.3 / 52.5\\
 & 5 & 0 & 80.7 / 40.2 & 72.2 / 41.1 & 77.2 / 40.2\\
 & 5 & 1 & 79.6 / 41.0 & 71.4 / 41.9 & 75.0 / 41.1\\
 & 10 & 0 & 88.6 / 41.5 & 81.4 / 41.1 & 83.7 / 41.4\\
\bottomrule
\end{tabular}
\end{center}
\end{table}

On the class constraints the equal step is lower in all eight runs, with FID$_{\mathrm{c}}$ within
$1.6$ of the standard rule in seven of them. The loss is $7.2$ to $8.5$ points on ship at
$\gamma=5$ and $10$, where the Wilson $95\%$ intervals of the two rules do not overlap, and $1.2$
to $4.4$ points on cat, in the same direction in every run. The test on a dependent conjunction
is in Appendix~\ref{app:depcomp} and the test on a molecular constraint in
Appendix~\ref{app:qm9alloc}.

\subsection{The normalized rule at small scales on CIFAR-10}
\label{app:normfront}

Section~\ref{sec:cifar} compares the rules at equal $\gamma$, where the normalized step is the
larger one. Table~\ref{tab:normfront} extends the normalized rule down to $\gamma=0.1$ so that the
two can be compared at equal satisfaction.

\begin{table}[H]
\caption{The matched classifier under the normalized rule (CG-norm) and ours over the guidance
scale: satisfaction (\%) / FID$_{\mathrm{c}}$, sampling seed $0$, $1{,}000$ samples, $1000$
steps; -- marks a setting that was not run.}
\label{tab:normfront}
\begin{center}
\small
\begin{tabular}{rllll}
\toprule
 & \multicolumn{2}{c}{Red-pixel} & \multicolumn{2}{c}{Upper luminance}\\
$\gamma$ & CG-norm & Ours & CG-norm & Ours\\
\midrule
0.1 & 39.3 / 52.6 & -- & 29.5 / 66.0 & -- \\
0.25 & 50.2 / 51.0 & -- & 44.7 / 56.3 & -- \\
0.5 & 74.4 / 53.3 & 55.4 / 49.7 & 69.8 / 45.1 & 55.8 / 54.8\\
1 & 95.1 / 61.7 & 80.3 / 49.4 & 96.8 / 43.2 & 90.9 / 45.0\\
2 & 100.0 / 108.1 & 97.7 / 50.6 & 99.9 / 62.3 & 99.2 / 39.5\\
5 & -- & 100.0 / 58.2 & 100.0 / 80.3 & 100.0 / 43.4\\
\bottomrule
\end{tabular}
\end{center}
\end{table}

Compared along the curves, each normalized point has a point of ours with satisfaction at
least as high and lower FID$_{\mathrm{c}}$. Three-seed means at $\gamma\ge2$ are in
Table~\ref{tab:cifarfull}.

\subsection{Composition of strongly dependent constraints}
\label{app:depcomp}

Red-pixel and low saturation are negatively dependent on CIFAR-10: $26.75\%$ and $25.03\%$ of
training images satisfy each, and $1.28\%$ satisfy both, against $6.69\%$ under independence.
Table~\ref{tab:depcomp} guides with $\log h_{\mathrm{red}}+\log h_{\mathrm{sat}}$ from the two
separately trained heads of Section~\ref{sec:cifar} ($1000$ steps).

\begin{table}[H]
\caption{Composition of red-pixel and low saturation: joint and marginal satisfaction (\%) and
the product of the two guided marginals; mean over $3$ sampling seeds $\times$ $1{,}000$ samples
($\pm$ standard deviation for the joint rate) at $\gamma=5$ and $10$, seed $0$ otherwise.
FID$_{\mathrm{c}}$ is against the $140$ test images that satisfy both.}
\label{tab:depcomp}
\begin{center}
\small
\begin{tabular}{rrrrrr}
\toprule
$\gamma$ & Joint & Red-pixel & Low saturation & Product & FID$_{\mathrm{c}}$\\
\midrule
0 & 0.5 & 31.4 & 15.1 & 4.7 & 160.64\\
2 & 30.4 & 84.2 & 39.5 & 33.3 & 124.91\\
5 & 62.5 $\pm$ 2.0 & 97.0 & 64.0 & 62.1 & 118.65\\
10 & 83.4 $\pm$ 0.3 & 97.6 & 83.9 & 81.9 & 131.05\\
\bottomrule
\end{tabular}
\end{center}
\end{table}

The joint rate rises from $0.5\%$ to $83.4\%$ and, from $\gamma=5$, is at or above the product of
the guided marginals in every seed. This is one pair of constraints; as stated in
Section~\ref{sec:method}, summed log-gradients are exact only under conditional independence, and
the result does not establish that arbitrary constraints compose.

\emph{Matched-budget test on the conjunction.} Table~\ref{tab:depalloc} applies the test of
Section~\ref{sec:alloc} to this conjunction. The direction is the gradient of
$\log h_{\mathrm{red}}+\log h_{\mathrm{sat}}$, $\bar m(t)$ is measured along the standard-rule
trajectories of the same seed, and the $1/\hat h$ rule uses
$\hat h=h_{\mathrm{red}}h_{\mathrm{sat}}$ floored at $10^{-3}$.

\begin{table}[H]
\caption{The test of Table~\ref{tab:alloc} on the conjunction of red-pixel and low saturation:
joint satisfaction (\%, mean $\pm$ standard deviation) / mean FID$_{\mathrm{c}}$ over $3$ sampling
seeds $\times$ $1{,}000$ samples.}
\label{tab:depalloc}
\begin{center}
\small
\begin{tabular}{rlll}
\toprule
$\gamma$ & Standard & Equal & $1/\hat h$\\
\midrule
5 & 62.5 $\pm$ 2.0 / 118.6 & 57.3 $\pm$ 2.2 / 118.8 & 61.6 $\pm$ 2.0 / 118.5\\
10 & 83.4 $\pm$ 0.3 / 131.1 & 76.1 $\pm$ 0.4 / 131.5 & 80.4 $\pm$ 1.1 / 133.6\\
\bottomrule
\end{tabular}
\end{center}
\end{table}

The equal step is lower in all six seed-level comparisons, by $5.2$ points on average at
$\gamma=5$ and $7.3$ at $\gamma=10$, at FID$_{\mathrm{c}}$ within $0.5$ of the standard rule; the
$1/\hat h$ weighting recovers $82\%$ and $59\%$ of the mean loss.

\subsection{The matched-budget test on a molecular constraint}
\label{app:qm9alloc}

Table~\ref{tab:qm9alloc} applies the test of Section~\ref{sec:alloc} to the Lipinski-type filter
of Section~\ref{sec:mol}, with the head, sampler and metrics of Appendix~\ref{app:qm9details}
(Euler--Maruyama, $1000$ steps, $1{,}000$ molecules per run). The guidance term acts on the node
features, $\bar m(t)$ is the mean norm of $\nabla\log h_\theta$ along the standard-rule trajectories
of the same seed, and $\hat h$ is floored at $10^{-3}$.

\begin{table}[H]
\caption{The test of Table~\ref{tab:alloc} on the QM9 Lipinski-type filter: satisfaction (\%, mean
$\pm$ standard deviation) and mean uniqueness / novelty (\%) over $3$ sampling seeds $\times$
$1{,}000$ molecules. Validity is $100\%$ in every run.}
\label{tab:qm9alloc}
\begin{center}
\small
\begin{tabular}{rlll}
\toprule
$\gamma$ & Standard & Equal & $1/\hat h$\\
\midrule
2 & 87.4 $\pm$ 0.5 (93.8 / 95.4) & 86.3 $\pm$ 1.2 (93.1 / 96.3) & 87.8 $\pm$ 0.7 (93.2 / 96.1)\\
5 & 93.7 $\pm$ 0.4 (89.5 / 94.4) & 91.0 $\pm$ 0.7 (89.5 / 95.3) & 93.5 $\pm$ 0.9 (89.2 / 94.7)\\
\bottomrule
\end{tabular}
\end{center}
\end{table}

The equal step is lower in all six seed-level comparisons. At $\gamma=5$ the loss is $2.7$ points
on average, between $1.5$ and $3.3$ points across the three seeds, and the $1/\hat h$ weighting
recovers over $90\%$ of it. At $\gamma=2$ the loss is $1.1$ points, in the same direction in all
three seeds. Validity is $100\%$ under all three rules, and uniqueness and novelty differ by less
than one point between them, so the loss in satisfaction is not offset by a gain in validity,
uniqueness or novelty.

\end{document}